\documentclass{article}

     \usepackage[preprint]{neurips_2024}

\usepackage{caption}
\usepackage[utf8]{inputenc} % allow utf-8 input
\usepackage[T1]{fontenc}    % use 8-bit T1 fonts
\usepackage{hyperref}       % hyperlinks
\usepackage{url}            % simple URL typesetting
\usepackage{booktabs}       % professional-quality tables
\usepackage{amsfonts}       % blackboard math symbols
\usepackage{nicefrac}       % compact symbols for 1/2, etc.
\usepackage{microtype}      % microtypography
\usepackage{xcolor}         % colors
\usepackage{graphicx}
\usepackage{multirow}
\usepackage{tikz}
\usepackage[edges]{forest}
\definecolor{hidden-draw}{RGB}{20,68,106}
\definecolor{hidden-pink}{RGB}{255,245,247}
\usepackage{amsmath}
\usepackage{setspace}
\usepackage{cleveref} 
\usepackage{makecell}
\usepackage{colortbl}
\usepackage{float}
\usepackage{longtable}
\usepackage{tabularx}

\title{A Survey on Benchmarks of Multimodal Large Language Models}

\author{%
      Jian Li\textsuperscript{1}\thanks{Corresponding author, <swordli@tencent.com>.}, \
      Weiheng Lu\textsuperscript{2}, \
      Hao Fei\textsuperscript{3}, \
      Meng Luo\textsuperscript{3}, \
      Ming Dai\textsuperscript{4}, \
      Min Xia\textsuperscript{2}, \
      \\
    \textbf{
      Yizhang Jin\textsuperscript{1}, \
      Zhenye Gan\textsuperscript{1}, \
      Ding Qi\textsuperscript{1}, \
      Chaoyou Fu\textsuperscript{5}, \
      Ying Tai\textsuperscript{5}, \
    } \\
    \textbf{
      Wankou Yang\textsuperscript{4}, \
      Yabiao Wang\textsuperscript{1}, \
      Chengjie Wang\textsuperscript{1}
    }
      \\
      \\
      \textsuperscript{1}Tencent, \
      \textsuperscript{2}PKU, \
      \textsuperscript{3}NUS, \
      \textsuperscript{4}SEU, \
      \textsuperscript{5}NJU
}

\begin{document}

\maketitle

\begin{abstract}
Multimodal Large Language Models (MLLMs) are gaining increasing popularity in both academia and industry due to their remarkable performance in various applications such as visual question answering, visual perception, understanding, and reasoning. Over the past few years, significant efforts have been made to examine MLLMs from multiple perspectives. This paper presents a comprehensive review of \textbf{200 benchmarks} and evaluation for MLLMs, focusing on (1)perception and understanding, (2)cognition and reasoning, (3)specific domains, (4)key capabilities, and (5)other modalities. Finally, we discuss the limitations of the current evaluation methods for MLLMs and explore promising future directions. Our key argument is that evaluation should be regarded as a crucial discipline to support the development of MLLMs better. For more details, please visit our GitHub repository: https://github.com/swordlidev/Evaluation-Multimodal-LLMs-Survey.

\end{abstract}

%\footnotetext[1]{* Equal contribution.}
\footnotetext[1]{Equal contribution: Jian Li and Weiheng Lu.}

\section{Introduction}

 Large Language Models (LLM) have recently garnered substantial interest across academic and industrial domains. The impressive performance of LLMs such as GPT~\cite{ouyang2022instructGPT} has fueled optimism that they could represent a step towards Artificial General Intelligence (AGI) in this era. The remarkable abilities of LLM have inspired efforts to integrate them with other modality-based models to enhance multimodal competencies. Consequently, Multimodal Large Language Models (MLLMs)~\cite{jin2024emllm} have emerged, This concept is further supported by the extraordinary success of proprietary models like OpenAI's GPT-4V~\cite{achiam2023gpt4} and Google's Gemini\cite{team2023gemini}. In contrast to earlier models that were limited to solving specific tasks, MLLMs demonstrate exceptional performance across a variety of applications, including both general Visual Question Answering (VQA) tasks and domain-specific challenges.

A comprehensive and objective benchmark for evaluating MLLMs is essential for comparing and investigating the performance of various models, and it plays a crucial role in the success of MLLMs. First, evaluating MLLMs helps us better understand the strengths and weaknesses of MLLMs. For example, the SEED-Bench~\cite{li2023SEED-Bench} illustrates that current MLLMs show weaker abilities in understanding spatial relationships between objects while achieving relatively high performance on global image comprehension. Second, evaluations across various scenarios can offer valuable guidance for MLLM applications in fields such as medicine, industry, and autonomous driving. This, in turn, can inspire future designs and expand the scope of their capabilities. Third, the broad applicability of MLLMs underscores the importance of ensuring their robustness, safety, and reliability, especially in safety-sensitive sectors. Finally, it is significant to evaluate other user-friendly features of MLLMs including the ability to handle long contexts and accurately follow instructions. Therefore, we aim to raise awareness in the community of the importance of MLLM evaluations by reviewing the current evaluation protocols.

Recently, numerous research efforts have focused on evaluating MLLMs from various perspectives, covering factors such as perception, understanding, cognition, and reasoning. Additionally, other MLLM capabilities have been tested, including robustness, trustworthiness, specialized applications, and different modalities. Despite these efforts, a comprehensive overview that captures the full scope of these evaluations is still lacking.

\begin{figure}[!t]
\centering
\includegraphics[width=0.95\linewidth]{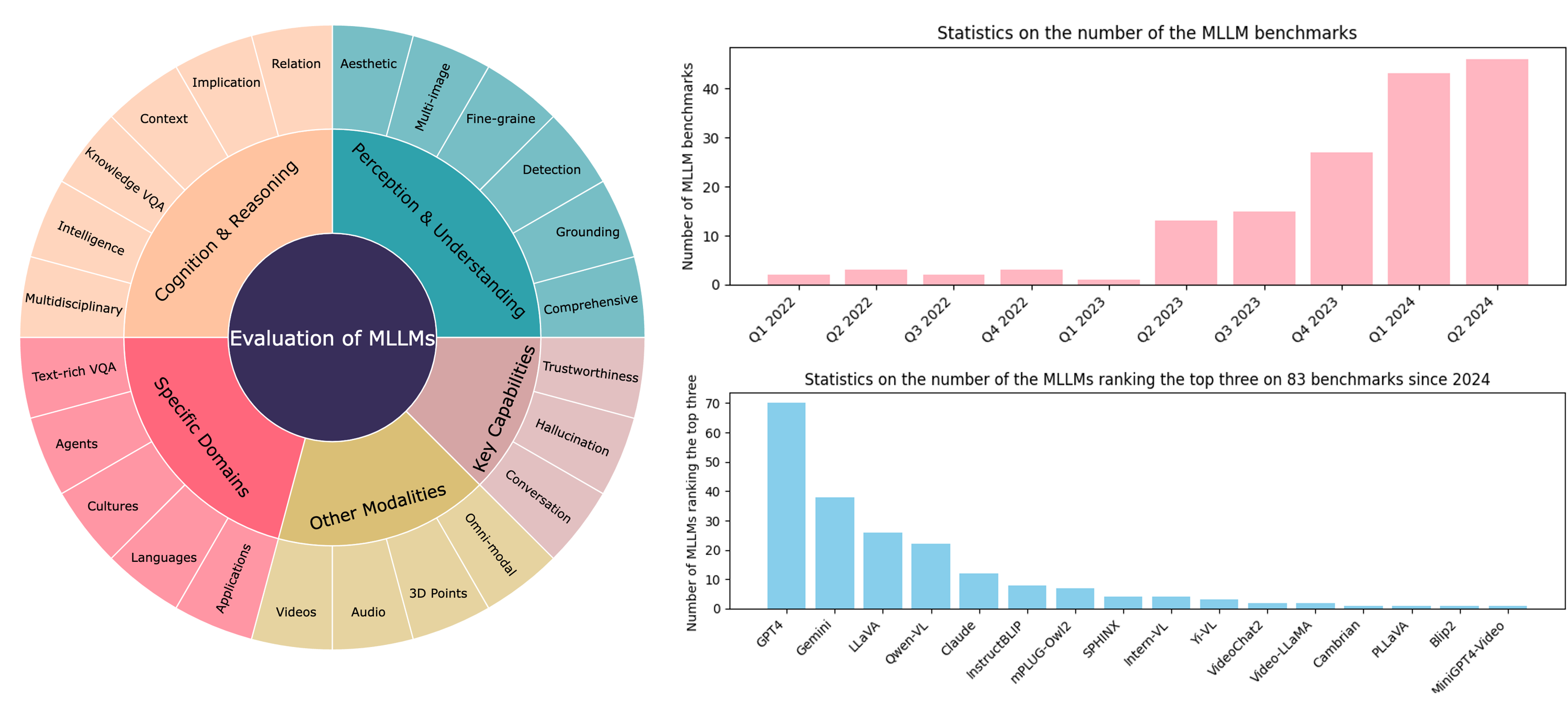}
\caption{(left) Taxonomy of this benchmarks survey, Our survey covers 5 key domains and 20-30 sub-class types, (Right up) Trend of MLLMs evaluation papers over time, (Right down) The statistics on the number of the top3 MLLMs on 83 benchmarks since 2024.}
\label{fig_statistic}
\end{figure}

\tikzstyle{my-box}=[
    rectangle,
    draw=hidden-draw,
    rounded corners,
    text opacity=1,
    minimum height=1.5em,
    minimum width=5em,
    inner sep=2pt,
    align=center,
    fill opacity=.5,
    line width=0.8pt,
]
\tikzstyle{leaf}=[my-box, minimum height=1.5em,
    fill=hidden-pink!80, text=black, align=left,font=\normalsize,
    inner xsep=2pt,
    inner ysep=4pt,
    line width=0.8pt,
]

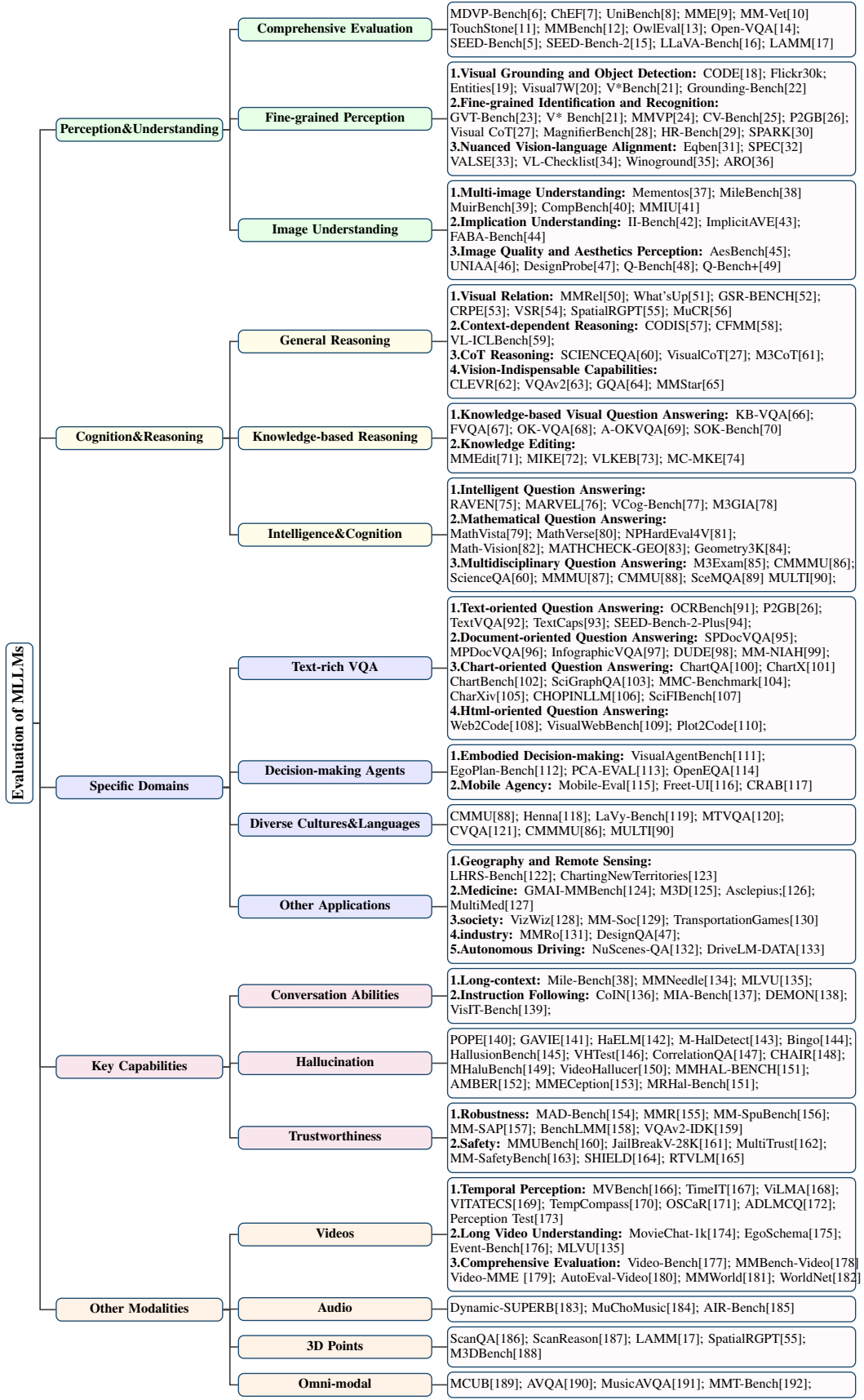
\begin{figure*}[]
    \centering
    \resizebox{\textwidth}{!}{
        \begin{forest}
            forked edges,
            for tree={
                grow=east,
                reversed=true,
                anchor=base west,
                parent anchor=east,
                child anchor=west,
                base=center,
                font=\large,
                rectangle,
                draw=hidden-draw,
                rounded corners,
                align=left,
                text centered,
                minimum width=4em,
                edge+={darkgray, line width=1pt},
                s sep=3pt,
                inner xsep=2pt,
                inner ysep=3pt,
                line width=0.8pt,
                ver/.style={rotate=90, child anchor=north, parent anchor=south, anchor=center},
            },
            where level=1{text width=12em,font=\normalsize,}{},
            where level=2{text width=14em,font=\normalsize,}{},
            where level=3{text width=16em,font=\normalsize,}{},
            where level=4{text width=35em,font=\normalsize,}{},
            where level=5{text width=18em,font=\normalsize,}{},
            [
                \textbf{Evaluation of MLLMs}, ver
                [
                        \textbf{Perception\&Understanding},
                        fill=green!10
                        [
                            \textbf{Comprehensive Evaluation}, fill=green!10
                            [
                                MDVP-Bench\cite{lin2024MDVP-Bench};
                                ChEF\cite{shi2023ChEF};
                                UniBench\cite{altahan2024UniBench};
                                MME\cite{MME};
                                MM-Vet\cite{yu2023MM-Vet}\\
                                TouchStone\cite{bai2023TouchStone}; 
                                MMBench\cite{liu2024MMBench};
                                OwlEval\cite{ye2024OwlEval};
                                Open-VQA\cite{zeng2023Open-VQA};\\
                                SEED-Bench\cite{li2023SEED-Bench};
                                SEED-Bench-2\cite{li2023SEED-Bench-2};
                                LLaVA-Bench\cite{liu2023LLaVA-Bench};
                                LAMM\cite{yin2023LAMM}\\
                                ,leaf, text width=30em
                            ]
                        ]
                        [
                            \textbf{Fine-grained Perception}, fill=green!10
                            [
                            \textbf{1.Visual Grounding and Object Detection:}
                                CODE\cite{zang2023CODE};
                                Flickr30k;\\ Entities\cite{plummer2016Flickr30kEntities}; 
                                Visual7W\cite{zhu2016Visual7W};
                                V*Bench\cite{wu2023V*Bench};
                                Grounding-Bench\cite{zhang2023Grounding-Bench}\\
                            \textbf{2.Fine-grained Identification and Recognition:} \\
                                GVT-Bench\cite{wang2023GVT-Bench};
                                V* Bench\cite{wu2023V*Bench};
                                MMVP\cite{tong2024MMVP};
                                CV-Bench\cite{tong2024CV-Bench};
                                P2GB\cite{chen2024P2GB};\\
                                Visual CoT\cite{shao2024VisualCoT};
                                MagnifierBench\cite{li2023MagnifierBench};
                                HR-Bench\cite{wang2024HR-Bench};
                                SPARK\cite{yu2024SPARK}\\
                            \textbf{3.Nuanced Vision-language Alignment:} 
                                Eqben\cite{wang2023Eqben};
                                SPEC\cite{peng2024SPEC}\\
                                VALSE\cite{VALSE};
                                VL-Checklist\cite{zhao2023VL-CheckList};
                                Winoground\cite{thrush2022Winoground};
                                ARO\cite{yuksekgonul2023ARO}\\
                            ,leaf, text width=30em
                            ]
                        ]
                        [
                            \textbf{Image Understanding}, fill=green!10
                              [
                                \textbf{1.Multi-image Understanding:} 
                                Mementos\cite{wang2024Mementos};
                                MileBench\cite{song2024MileBench}\\
                                MuirBench\cite{wang2024MuirBench};
                                CompBench\cite{kil2024CompBench};
                                MMIU\cite{meng2024MMIU}\\
                                \textbf{2.Implication Understanding:} 
                                II-Bench\cite{liu2024II-Bench};
                                ImplicitAVE\cite{zou2024ImplicitAVE};\\
                                FABA-Bench\cite{li2024FABA-Bench}\\
                                \textbf{3.Image Quality and Aesthetics Perception:} 
                                AesBench\cite{huang2024AesBench}; \\
                                UNIAA\cite{zhou2024UNIAA};
                                DesignProbe\cite{doris2024DesignQA};
                                Q-Bench\cite{wu2024Q-Bench};
                                Q-Bench+\cite{zhang2024Q-Bench+}\\
                                , leaf, text width=30em
                              ]
                        ]
                    ]
                     [
                        \textbf{Cognition\&Reasoning}, fill=yellow!10
                        [
                            \textbf{General Reasoning}, fill=yellow!10
                            [
                            \textbf{1.Visual Relation:} 
                            MMRel\cite{nie2024MMRel};
                            What'sUp\cite{kamath2023what'sup};
                            GSR-BENCH\cite{rajabi2024GSR-BENCH};\\
                            CRPE\cite{wang2024CRPE};
                            VSR\cite{VSR};
                            SpatialRGPT\cite{cheng2024SpatialRGPT};
                            MuCR\cite{li2024MuCR}\\
                            \textbf{2.Context-dependent Reasoning:} 
                            CODIS\cite{luo2024CODIS};
                            CFMM\cite{li2024CFMM};\\
                            VL-ICLBench\cite{zong2024VL-ICLBench};\\
                            \textbf{3.CoT Reasoning:} 
                            SCIENCEQA\cite{lu2022ScienceQA};
                            VisualCoT\cite{shao2024VisualCoT};
                            M3CoT\cite{chen2024M3CoT};\\
                            \textbf{4.Vision-Indispensable Capabilities:} \\
                            CLEVR\cite{johnson2016CLEVR};
                            VQAv2\cite{goyal2017VQAv2};
                            GQA\cite{hudson2019GQA};
                            MMStar\cite{chen2024MMStar}
                             , leaf, text width=30em
                            ]
                        ]
                        [
                            \textbf{Knowledge-based Reasoning}, fill=yellow!10
                            [
                            \textbf{1.Knowledge-based Visual Question Answering:} 
                            KB-VQA\cite{wang2015KBVQA};\\
                            FVQA\cite{wang2017FVQA};
                            OK-VQA\cite{marino2019OK-VQA};
                            A-OKVQA\cite{schwenk2022A-OKVQA};
                            SOK-Bench\cite{wang2024SOK-Bench}\\
                            \textbf{2.Knowledge Editing:} \\
                            MMEdit\cite{cheng2024MMEdit};
                            MIKE\cite{li2024MIKE};
                            VLKEB\cite{huang2024VLKEB};
                            MC-MKE\cite{MC-MKE}
                            , leaf, text width=30em
                            ]
                        ][
                            \textbf{Intelligence\&Cognition}, fill=yellow!10
                            [
                            \textbf{1.Intelligent Question Answering:} \\
                            RAVEN\cite{zhang2019RAVEN};
                            MARVEL\cite{jiang2024MARVEL};
                            VCog-Bench\cite{cao2024VCog-Bench};
                            M3GIA\cite{song2024M3GIA}\\
                            \textbf{2.Mathematical Question Answering:}\\
                            MathVista\cite{lu2024MathVista};
                            MathVerse\cite{MathVerse};
                	      NPHardEval4V\cite{fan2024NPHardEval4V};\\
                            Math-Vision\cite{wang2024Math-Vision};
                            MATHCHECK-GEO\cite{zhou2024MATHCHECK};
                            Geometry3K\cite{lu2021Geometry3K};\\
                            \textbf{3.Multidisciplinary Question Answering:} 
                            M3Exam\cite{zhang2023M3Exam};
                            CMMMU\cite{zhang2024CMMMU};\\
                            ScienceQA\cite{lu2022ScienceQA};
                            MMMU\cite{yue2024MMMU};
                            CMMU\cite{he2024CMMU};
                            SceMQA\cite{liang2024SceMQA}
                            MULTI\cite{zhu2024MULTI};
                            , leaf, text width=30em
                            ]
                        ]
                    ]
                    [
                        \textbf{Specific Domains}, fill=blue!10
                        [
                            \textbf{Text-rich VQA}, fill=blue!10
                            [\textbf{1.Text-oriented Question Answering:}
                            OCRBench\cite{liu2024OCRBench};
                            P2GB\cite{chen2024P2GB};\\
                            TextVQA\cite{singh2019TextVQA};
                            TextCaps\cite{sidorov2020TextCaps};
                            SEED-Bench-2-Plus\cite{li2024SEED-Bench-2-Plus};\\
                            \textbf{2.Document-oriented Question Answering:} 
                            SPDocVQA\cite{mathew2021SPDocVQA};\\ 
                            MPDocVQA\cite{tito2023MPDocVQA};
                            InfographicVQA\cite{mathew2021infographicvqa};
                            DUDE\cite{vanlandeghem2023DUDE};
                            MM-NIAH\cite{wang2024MM-NIAH};\\
                            \textbf{3.Chart-oriented Question Answering:} 
                            ChartQA\cite{masry2022ChartQA};
                            ChartX\cite{xia2024ChartX}\\
                            ChartBench\cite{xu2024ChartBench};
                            SciGraphQA\cite{li2023SciGraphQA};
                            MMC-Benchmark\cite{liu2024MMC-Benchmark};\\
                            CharXiv\cite{wang2024CharXiv};
                            CHOPINLLM\cite{fan2024CHOPINLLM};
                            SciFIBench\cite{roberts2024SciFIBench}\\
                            \textbf{4.Html-oriented Question Answering:}\\
                            Web2Code\cite{yun2024Web2Code};
                            VisualWebBench\cite{liu2024VisualWebBench};
                            Plot2Code\cite{wu2024Plot2Code};\\
                            , leaf, text width=30em
                            ]   
                        ]
                        [
                            \textbf{Decision-making Agents}, fill=blue!10
                            [\textbf{1.Embodied Decision-making:}
                            VisualAgentBench\cite{liu2024VisualAgentBench};\\
                            EgoPlan-Bench\cite{chen2024EgoPlan-Bench};
                            PCA-EVAL\cite{chen2023PCA-EVAL};
                            OpenEQA\cite{OpenEQA2023}\\
                            \textbf{2.Mobile Agency:} 
                            Mobile-Eval\cite{wang2024Mobile-Eval};
                            Freet-UI\cite{you2024Ferret-UI};
                            CRAB\cite{xu2024CRAB}\\
                            , leaf, text width=30em
                            ]
                        ]
                        [
                            \textbf{Diverse Cultures\&Languages}, fill=blue!10
                            [CMMU\cite{he2024CMMU};
                            Henna\cite{alwajih2024Henna};
                            LaVy-Bench\cite{tran2024LaVy};
                            MTVQA\cite{tang2024MTVQA};\\
                            CVQA\cite{romero2024CVQA};
                            CMMMU\cite{zhang2024CMMMU};
                            MULTI\cite{zhu2024MULTI}
                            , leaf, text width=30em
                            ]
                        ]
                        [
                            \textbf{Other Applications}, fill=blue!10
                            [
                            \textbf{1.Geography and Remote Sensing:}\\
                            LHRS-Bench\cite{muhtar2024LHRS-Bench};
                            ChartingNewTerritories\cite{roberts2024ChartingNewTerritories}\\
                            \textbf{2.Medicine:}
                            GMAI-MMBench\cite{chen2024GMAI-MMBench};
                            M3D\cite{bai2024M3D};
                            Asclepius;\cite{wang2024Asclepius}; \\
                            MultiMed\cite{mo2024MultiMed}\\
                            \textbf{3.society:}
                            VizWiz\cite{gurari2018VizWiz};
                            MM-Soc\cite{jin2024MM-Soc};
                            TransportationGames\cite{zhang2024TransportationGames}\\
                            \textbf{4.industry:}
                            MMRo\cite{li2024MMRo};
                            DesignQA\cite{doris2024DesignQA};\\
                            \textbf{5.Autonomous Driving:}
                            NuScenes-QA\cite{qian2024NuScenes-QA};
                            DriveLM-DATA\cite{sima2024DriveLM-Data}
                            , leaf, text width=30em
                            ]
                        ]
                    ]
                    [
                        \textbf{Key Capabilities}, 
                        fill=purple!10
                        [
                            \textbf{Conversation Abilities}, fill=purple!10
                            [
                            \textbf{1.Long-context:} 
                            Mile-Bench\cite{song2024MileBench};
                            MMNeedle\cite{wang2024MMNeedle};
                            MLVU\cite{zhou2024MLVU};\\
                            \textbf{2.Instruction Following:}
                            CoIN\cite{chen2024CoIN};
                            MIA-Bench\cite{qian2024MIA-Bench};
                            DEMON\cite{wang2023DEMON};\\
                            VisIT-Bench\cite{bitton2023VisIT-Bench};
                            , leaf, text width=30em
                            ]   
                        ]
                        [
                            \textbf{Hallucination}, fill=purple!10
                            [
                            POPE\cite{li2023POPE};
                            GAVIE\cite{liu2024GAVIE};
                            HaELM\cite{wang2023HaELM};
                            M-HalDetect\cite{gunjal2024M-HalDetect};
                            Bingo\cite{cui2023Bingo}; \\
                            HallusionBench\cite{guan2024HallusionBench};
                            VHTest\cite{huang2024VHTest};
                            CorrelationQA\cite{han2024CorrelationQA};
                            CHAIR\cite{rohrbach2019CHAIR};\\
                            MHaluBench\cite{chen2024MHaluBench};
                	      VideoHallucer\cite{wang2024VideoHallucer};
                            MMHAL-BENCH\cite{sun2023MMHAL-BENCH};\\
                            AMBER\cite{wang2024AMBER};
                	      MMECeption\cite{cao2024MMECeption};
                            MRHal-Bench\cite{sun2023MMHAL-BENCH};
                            , leaf, text width=30em
                            ]
                        ]
                        [
                            \textbf{Trustworthiness}, fill=purple!10
                            [
                            \textbf{1.Robustness:}
                            MAD-Bench\cite{qian2024MAD-Bench};
                            MMR\cite{liu2024MMR};
                	      MM-SpuBench\cite{ye2024MM-SpuBench};\\
                            MM-SAP\cite{wang2024MM-SAP};
                	      BenchLMM\cite{cai2023benchlmm};
                            VQAv2-IDK\cite{cha2024VQAv2-IDK}\\
                            \textbf{2.Safety:}
                            MMUBench\cite{li2024MMUBench};
                            JailBreakV-28K\cite{luo2024JailBreakV-28K};
                            MultiTrust\cite{zhang2024MultiTrust};\\
                	      MM-SafetyBench\cite{liu2024MM-SafetyBench};
                            SHIELD\cite{shi2024SHIELD};
                	      RTVLM\cite{li2024RTVLM}\\
                            , leaf, text width=30em
                            ]
                        ]
                    ]
                [
                    \textbf{Other Modalities}, fill=orange!10
                    [
                        \textbf{Videos},  fill=orange!10
                        [
                        \textbf{1.Temporal Perception:}
                        MVBench\cite{li2024MVBench};
                        TimeIT\cite{Ren2023TimeIT};
                        ViLMA\cite{kesen2023ViLMA};\\
                        VITATECS\cite{li2023VITATECS};
                        TempCompass\cite{liu2024TempCompass};
                        OSCaR\cite{nguyen2024OSCaR};
                        ADLMCQ\cite{chakraborty2024ADLMCQ};\\
                        Perception Test\cite{pătrăucean2023PerceptionTest}\\
                        \textbf{2.Long Video Understanding:}
                        MovieChat-1k\cite{song2024MovieChat};
                        EgoSchema\cite{mangalam2023EgoSchema};\\
                        Event-Bench\cite{du2024Event-Bench};
                        MLVU\cite{zhou2024MLVU}\\
                        \textbf{3.Comprehensive Evaluation:}
                        Video-Bench\cite{ning2023Video-Bench};
                        MMBench-Video\cite{fang2024MMBench-Video} \\
                        Video-MME \cite{fu2024Video-MME};
                        AutoEval-Video\cite{chen2024AutoEval-Video};
                        MMWorld\cite{he2024MMWorld};
                        WorldNet\cite{ge2024WorldNet-Crafted}
                        , leaf, text width=30em
                        ]
                    ]
                    [
                        \textbf{Audio},  fill=orange!10
                        [
                        Dynamic-SUPERB\cite{huang2024Dynamic-SUPERB};
                        MuChoMusic\cite{weck2024MuChoMusic};
                        AIR-Bench\cite{yang2024AIR-Bench}
                        , leaf, text width=30em
                        ]
                    ]
                    [
                        \textbf{3D Points},  fill=orange!10
                        [
                        ScanQA\cite{azuma2022ScanQA};
                        ScanReason\cite{zhu2024ScanReason};
                        LAMM\cite{yin2023LAMM};
                        SpatialRGPT\cite{cheng2024SpatialRGPT};\\
                        M3DBench\cite{li2023M3DBench}
                        , leaf, text width=30em
                        ]
                    ]
                    [
                     \textbf{Omni-modal}, fill=orange!10
                      [
                        MCUB\cite{chen2024MCUB}; 
                        AVQA\cite{AVQA};
                        MusicAVQA\cite{li2022MusicAVQA};
                        MMT-Bench\cite{ying2024MMT-Bench};\\
                        ,leaf, text width=30em
                     ]
                    ]
                ]
            ]
        \end{forest}
      }
    \vspace{-0mm}
    \caption{Taxonomy of benchmarks for Multimodal Large Language Models}
    \label{fig: MLLMs evaluations structure}
    \vspace{0mm}
\end{figure*}

In this survey, we aim to provide a comprehensive overview of recent advancements in the rapidly evolving field of MLLM evaluation. As illustrated in Figure~\ref{fig_statistic}, our survey covers five key areas of MLLM evaluation, encompassing 20-30 detailed categories. The figure also illustrates the trend of MLLM evaluation papers over time, revealing a rapid increase in the number of publications. This growth indicates that the research area has garnered widespread attention. Furthermore, we provide statistics on the performance of the top three MLLMs across 83 benchmarks since 2024. The data indicates that OpenAI's GPT-4 and Google's Gemini exhibit superior performance and have attracted significant academic attention. As depicted in Figure.~\ref{fig: MLLMs evaluations structure}, we survey 200 benchmarks and organize the literature in a taxonomy consisting of five primary categories, encompassing various aspects, focusing on \textbf{perception and understanding; cognition and reasoning; specific domains; key capabilities and other modalities.} 

\begin{itemize}
    \item[\textbf{•}] \textbf{Perception and Understanding} refer to the ability to receive and extract features from multimodal data and perform cross-modal analysis. Evaluating MLLMs' perception and understanding capabilities includes assessing whether MLLMs can perceive visual representations, identify visual details, grasp the meaning and emotions conveyed by images, and respond to related questions correctly. These abilities are the cornerstone of MLLMs, enabling them to perform a wide range of tasks and applications.

    \item[\textbf{•}] \textbf{Cognition and Reasoning} encompass the model’s capacity for advanced processing and complex inference beyond basic perception and understanding. Cognitive abilities involve processing and manipulating information to transform it into knowledge, while reasoning abilities focus on drawing logical conclusions and solving problems. Strong cognitive and reasoning abilities enable MLLMs to perform effective logical inference in complex tasks.

    \item[\textbf{•}] \textbf{Specific domains} focus on MLLMs' capabilities in particular tasks and applications, such as their ability to process text-rich visual information and perform agent-based decision-making tasks in real-world scenarios. The discussion then extends to the evaluation of their performance in specialized domains such as medicine, autonomous driving, and industry.
    % focusing on their performance across a range of specialized tasks. The discussion extends to the practical applications of MLLMs, highlighting their impact on operational efficiency in various sectors such as medicine, industry, and autonomous driving

    \item[\textbf{•}] \textbf{Key capabilities} significantly impact the performance and user experience of MLLMs, including managing complex dialogues, accurately following instructions, and avoiding hallucinations while maintaining trustworthiness. These capabilities are essential for ensuring that MLLMs operate effectively across diverse real-world applications and adapt to various practical scenarios.
    % focus on important aspects related to user experience during usage. These include dialogue capabilities, specifically handling long contexts and accurately following instructions, as well as the model’s level of hallucination and trustworthiness.

    \item[\textbf{•}] \textbf{Other modalities} include video, audio, and 3D point clouds, which also contain rich and diverse information reflective of the real world. These modalities provide critical context and enhance the ability of MLLMs to understand complex scenarios. Evaluating MLLMs' ability to handle various modalities can help in understanding their performance across different types of data and tasks, ensuring they are suitable for complex real-world scenarios and challenging tasks.

\end{itemize}

\section{Preliminaries}

\begin{figure}[!t]
\centering
\includegraphics[width=0.95\linewidth]{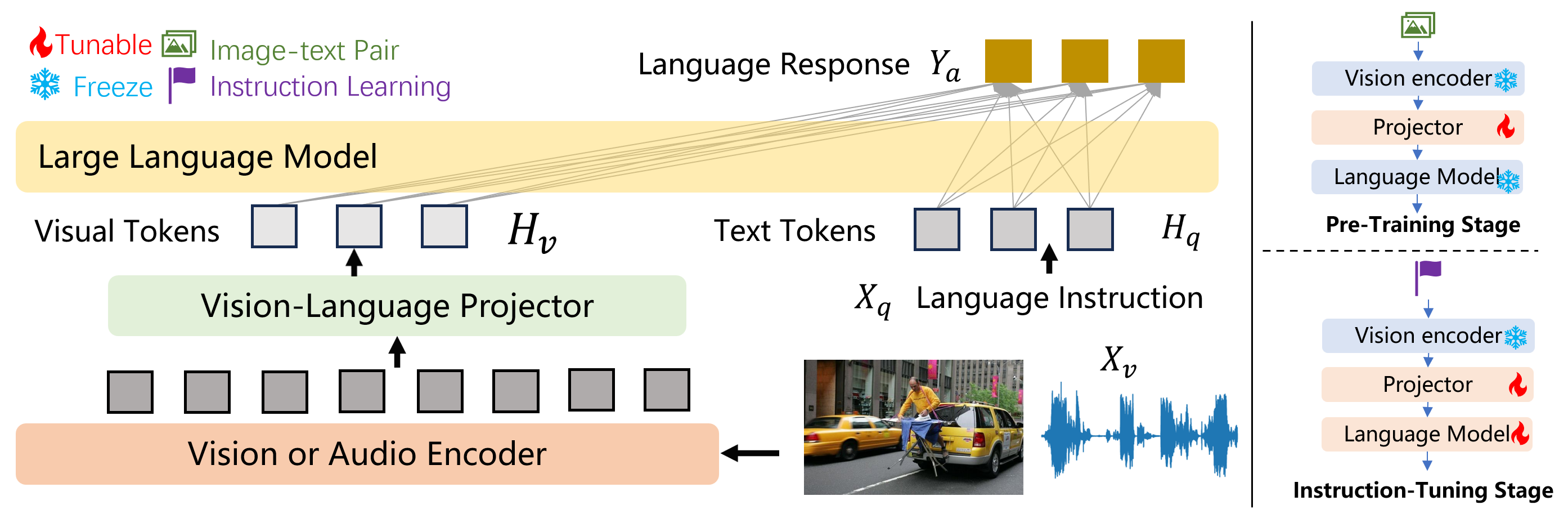}
\caption{The architectures and training process of MLLMs.}
\label{fig_architecture}
\end{figure}

Figure.~\ref{fig_statistic} compares several common MLLMs including GPT4\cite{achiam2023gpt4}, Gemini\cite{team2023gemini}, LLaVA\cite{liu2023llava}, Qwen-VL\cite{bai2023Qwen-VL}, Claude\cite{anthropic2024claude},
  InstructBLIP\cite{dai2024instructblip}, mPLUG-Owl2\cite{ye2023mplugowl2}, SPHINX\cite{lin2023SPHINX}, Intern-VL\cite{chen2023internvl}, Yi-VL\cite{ai2024Yi-VL}, VideoChat2\cite{li2023videochat}, Video-LLaMA\cite{zhang2023Video-LLaMA},
 Cambrian-1\cite{tong2024Cambrian}, PLLaVA\cite{xu2024PLLaVA}, Blip2\cite{li2023blip2}, MiniGPT4-Video\cite{ataallah2024MiniGPT4-Video}. The standard MLLM framework can be divided into three main modules: a visual encoder $g$ tasked with receiving and processing visual inputs, a pre-trained language model that manages the received multimodal signals and performs reasoning, and a visual-language projector $P$ which functions as a bridge to align the two modalities. A diagram of the architecture and training process is illustrated in Figure.~\ref{fig_architecture}. This figure outlines the base LLM, the vision encoder, and the projector, as well as the pretraining and instruction tuning.

\subsection{MLLM Architecture}

\paragraph{Vision Encoder} 
Taking the input image $X_v$ as input, the vision encoder compresses the original image into more compact patch features $Z_v$, as represented by the following formula:
\begin{equation}
Z_v = g(X_v).    
\end{equation} 

The encoder $g$ can also be an audio encoder for audio feature extraction or an encoder for other modalities. The common vision encoders are CLIP~\cite{radford2021clip}, SigLIP~\cite{zhai2023siglip} and DINO~\cite{caron2021dinov1,oquab2023dinov2}. 

\paragraph{Vision-Language Projector}
The task of the vision-language projector is to map the visual patch embeddings $Z_v$ into the text feature space:
\begin{equation}
    H_v = P(Z_v),
\end{equation}
where $H_v$ denotes the projected visual embeddings. The aligned visual features are used as prompts and inputted into the language model along with the text embeddings. Several works, such as Qformer in BLIP-2~\cite{li2023blip2}, design new projectors to reduce the number of visual tokens for efficiency.

\paragraph{Large Language Model}
The pre-trained Large language model serves as the core component of MLLMs, endowing it with many outstanding capabilities, such as zero-shot generalization, instruction following, and in-context learning. The LLM accepts input sequences containing multiple modalities and outputs corresponding text sequences. A text tokenizer is typically bundled with the LLM, mapping text prompts $X_q$ to the text tokens $H_q$. The text tokens $H_q$ and the visual tokens $H_v$ are concatenated as the input of the language model, which outputs the final response sequence $Y_a$ in an autoregressive manner:
\begin{equation}
    p(Y_a|H_v,H_q)=\prod^L_{i=1}p(y_i|H_v,H_q,y_{<i}),
\end{equation}
\label{eq1}
where $L$ denotes the length of $Y_a$. The parameter sizes of large language models (LLMs) range from 3 billion to tens of billions. Commonly used open-source LLMs include the Llama series\cite{touvron2023llama,touvron2023llama2,chiang2023vicuna,zhang2024tinyllama}, Phi~\cite{li2023phi,abdin2024phi3},Gemma~\cite{gemmateam2024gemma},Qwen~\cite{bai2023qwen}.

\subsection{MLLM Training}
The standard training process of MLLMs is a crucial factor that determines their performance on downstream tasks and their ability to handle diverse tasks. In this section, we provide an overview of various training methodologies, including pre-training, and instruction-tuning.

\paragraph{Pre-training} 
The pre-training(PT) stage focuses on aligning different modalities within the embedding space, enabling the language model to accept inputs from various modalities. This phase primarily involves large-scale text-paired data, predominantly in the form of image-caption pairs. An image-caption pair $(X, Y)$ is typically expanded into a single-turn conversation $(X_{instruct}, X_a)$, where $X_{instruct}$ contains an image $X_v$ and a randomly sampled question $X_q$ from a set of instructions asking the assistant to briefly describe the image, and $X_a$ is the original image description. Given such a conversation, the model is trained to autoregressively predict the image description. Consequently, following the "next token prediction" paradigm, we can compute the probability of predicting $X_a$ conditioned by $X_v$ and optimize it using a standard cross-entropy loss function:
\begin{equation}
    \mathop{\max}_{\theta} \sum_{i=1}^L \log p_\theta(x_i|X_v,X_{instruct},X_{a,<i}),
\end{equation}
where $L$ is the length of $X_a$ and $\theta$ denotes the trainable parameters. In order to better align different modalities of knowledge and avoid catastrophic forgetting during the PT stage, $\theta$ typically includes only a learnable modality interface, \textit{i.e.}, a vision-language projector. 

\paragraph{Instruction-tuning} 
Instruction-tuning(IT) aims to fine-tune the models on specific tasks by leveraging task-specific instructions. IT is typically conducted within the paradigm of Supervised Fine-Tuning (SFT). The IT datasets
are transformed into an instruction-based format, presented in the form of single-turn or multi-turn dialogue structures. Given an image $X_v$ and its caption, a conversation data $(X_q^1,X_a^1,\ldots,X_q^T,X_a^T)$ can be generated, where T is the total number of turns. 
Typically, we can organize the data into a sequence of instructions and responses, as outlined in \cite{liu2023llava}. With this multimodal instruction-following sequence, IT can be performed using the same auto-regressive training objective as in the PT stage. A common strategy involves maintaining the visual encoder weights fixed while continuing to update the PT weights of both the projector and the LLM during the IT process.

%\section{Evaluating MLLMs from Multiple Perspectives}
%\label{headings}
%In this chapter, we provide a comprehensive overview of the benchmarks used to evaluate Multimodal Large Language Models (MLLMs) across various key dimensions. We categorize these benchmarks into six major categories: perceptual and understanding capabilities, logical reasoning capabilities, domain-specific competencies, hallucination assessment, safety and robustness, and comprehensive capability evaluation. This analysis aims to offer a thorough understanding of the current standards and practices in MLLM evaluation, highlighting the strengths and limitations of existing approaches. The insights gained from this overview are intended to guide future research and development in the field, promoting the advancement of more effective and reliable MLLM systems.

\section{Perception and Understanding}
\label{sec:pu}
When evaluating the perception and understanding capabilities of MLLMs, we focus on benchmarks that assess the model's fundamental abilities in visual information processing. This includes evaluating the MLLMs’ accuracy in object identification and detection, understanding of scene context and object relationships, and ability to respond to questions about image content. Perception and understanding abilities are the cornerstone of MLLMs, enabling them to perform a wide range of tasks and applications. This section first introduces comprehensive evaluation benchmarks for MLLMs, and then separately discusses coarse-grained and fine-grained benchmarks for visual perception.

\subsection{Comprehensive Evaluation}
MLLMs rely on the powerful LLM to perform multimodal tasks, showing amazing emergent abilities in various studies. In order to fully match the flourish of MLLMs, Many comprehensive evaluation benchmarks are proposed.

To advance research on visual-related tasks, LLaVA-Bench \cite{liu2023LLaVA-Bench} and OwlEval \cite{ye2024OwlEval} were constructed to examine a variety of MLLM capabilities, but the quantity of these benchmarks was too small to fully reflect the performance of MLLMs. There was still an issue of lacking a more comprehensive evaluation with large-scale data.
Fu et al. \cite{MME} filled this gap by presenting the first comprehensive MLLM evaluation benchmark, MME, which measured both perception and cognition abilities across a range of subtasks.
Considering that subjective benchmarks like OwlEval relied on human labor for evaluations, which was not scalable and could exhibit significant bias, Liu et al. \cite{liu2024MMBench} proposed employing GPT-4 \cite{achiam2023gpt4} to match MLLM predictions and devised MMBench to robustly assess various abilities of MLLMs.
However, both the binary judgments used in MME and the multiple-choice answer selection employed by MMBench could not fully capture the complexity of open-ended real-world dialogues. In light of this concern, Open-VQA \cite{zeng2023Open-VQA} and TouchStone \cite{bai2023TouchStone} were proposed to support open-ended answers. Nevertheless, the relatively small scale of these benchmarks introduced instability in the evaluation statistics.
To overcome this limitation, SEED-Bench was introduced with annotations six times larger than existing benchmarks. It included a substantial number of multiple-choice questions covering various evaluation dimensions for both image and video modalities.
Additionally, SEED-Bench-2 categorized MLLMs' capabilities into hierarchical levels from L0 to L4 and served as a benchmark for evaluating MLLMs' hierarchical capabilities.
Despite the promising qualitative results regarding MLLMs’ capabilities, it remained unclear how to systematically evaluate complex multimodal tasks and what the relationships among the evaluated tasks were. Based on this consideration, MM-Vet \cite{yu2023MM-Vet} was devised to study integrated vision-language capabilities, allowing the evaluation to provide insights beyond overall model rankings.
Moreover, MDVP-Bench \cite{lin2024MDVP-Bench} and LAMM \cite{yin2023LAMM} were created to provide a comprehensive assessment of MLLMs' capabilities, particularly in understanding visual prompting instructions.
Furthermore, to provide a fair and systematic assessment of MLLMs’ performance across diverse multimodal tasks, ChEF \cite{shi2023ChEF} and UniBench \cite{altahan2024UniBench} were constructed as standardized and holistic evaluation frameworks, which helped to comprehensively understand the capabilities and limitations of MLLMs.

\subsection{Fine-grained Perception}
One indispensable cornerstone of MLLMs is the ability to perceive visible objects within scenes precisely. This includes evaluating MLLMs' capabilities in object detection and recognition, understanding details within local regions, and achieving accurate vision-language alignment. Such fine-grained perception is crucial for effective multimodal understanding and interaction.
% One indispensable cornerstone of MLLMs is the ability to perceive visible objects within scenes precisely, 

\textbf{Visual Grounding and Object Detection}: Object-level grounding and detection are critical steps in better perceiving images and solving image-related questions, providing a stronger link between QA pairs and images. 
Flickr30k Entities \cite{plummer2016Flickr30kEntities} and Visual7W \cite{zhu2016Visual7W} were early benchmarks for entity localization, focusing on the detailed grounding of specific phrases in image regions and QA tasks related to local areas. These benchmarks served as the foundation for subsequent developments in the field.
% including the detailed grounding of specific phrases in image regions and QA tasks related to local regions, serving as the foundation for subsequent benchmarks. 
However, they did not provide enough detailed contextual information about objects within the scene. Zang et al. \cite{zang2023CODE} investigated contextual object detection, which involved understanding visible objects within human-AI interactive contexts, and presented the CODE benchmark to facilitate research in this area. To evaluate MLLMs' ability in challenging scenarios where images contained abundant and complex information, Wu et al. \cite{wu2023V*Bench} introduced V*Bench, a benchmark focused on detailed visual grounding in high-resolution images. 
However, these benchmarks for visual grounding primarily used data with short captions and overlooked MLLMs' chat performance when asked to ground. To address this issue, Zhang et al.\cite{zhang2023Grounding-Bench} introduced a benchmark called Grounding-Bench, which allowed for combined evaluation of both grounding and chat capabilities.
% Flickr30k Entities \cite{plummer2016Flickr30kEntities} and Visual7W \cite{zhu2016Visual7W} were early benchmarks for the localization of entities, including the detailed grounding of specific phrases in image regions and QA tasks related to local regions, serving as the foundation for subsequent benchmarks. However, they didn't provide enough detailed contextual information about objects within the scene. Zang et al.\cite{zang2023CODE} were the first to investigate contextual object detection, which means understanding visible objects within human-AI interactive contexts, and presented the CODE benchmark to facilitate research in this area. To quantitatively evaluate MLLM’s ability in challenging scenarios where the image contains abundant and complex information, Wu et al. \cite{wu2023V*Bench} introduced V*Bench, a benchmark focused on detailed visual grounding on high-resolution images.

\textbf{Fine-grained Identification and Recognition:}Fine-grained identification and recognition requires MLLMs to identify and analyze detailed visual features.
% which is crucial for assessing whether the visual tokenizer is optimal for MLLMs. 
In order to determine what makes for a good visual tokenizer, GVT-bench\cite{wang2023GVT-Bench} was designed to evaluate MLLMs' fine-grained visual perception through object counting and multi-class identification. However, the data used in GVT-bench were constrained to relatively small resolutions. 
To address this limitation, MagnifierBench\cite{li2023MagnifierBench} and HR-Bench\cite{wang2024HR-Bench} extended the input resolution to a significantly larger range, thereby evaluating the ability of MLLMs to discern details in high-resolution images.
% To address this limitation, Li et al. \cite{li2023MagnifierBench} extended the input resolution to a much larger range and introduced MagnifierBench, which tested MLLMs' ability to discern the details of small objects in high-resolution images.
Considering that most open-source MLLMs adopted the Pretrained Contrastive Language-Image PreTraining (CLIP) model\cite{radford2021clip} as the visual encoder, Tong et al. \cite{tong2024MMVP} proposed the MMVP benchmark consisting of CLIP-blind pairs. They found that all tested models struggled with simple questions about visual details.
Given that existing benchmarks were insufficient in size and did not cover crucial visual elements such as depth and spatial awareness, CV-Bench\cite{tong2024CV-Bench} was created to evaluate the fundamental 2D and 3D visual understanding of MLLMs. 
Furthermore, Yu et al.\cite{yu2024SPARK} proposed a benchmark named SPARK, which focused on evaluating the ability of MLLMs to analyze images captured by various multi-vision sensors, including RGB, thermal, depth, and X-ray images.
% Given that existing benchmarks were insufficient in size and did not cover crucial visual elements such as depth and spatial awareness, CV-Bench \cite{tong2024CV-Bench} was created to evaluate the fundamental 2D and 3D visual understanding of MLLMs, using a significantly larger number of examples. 
Additionally, P2GB \cite{chen2024P2GB} was designed to assess fine-grained visual capabilities, especially in text-rich scenarios, requiring comprehensive recognition and understanding of image text content.
Moreover, VisualCoT \cite{shao2024VisualCoT} was constructed with visual chain-of-thought prompts, allowing MLLMs to focus on specific regions within a complete image, similar to how humans comprehend intricate visual information.
% In order to determine what makes for a good visual tokenizer, GVT-bench was designed to evaluate MLLMs' fine-grained visual perception through object counting and multi-class identification, but the data used in GVT-Bench were constrained to relatively small resolutions. Li et al.\cite{li2023MagnifierBench}extended the input resolution to a much larger range to fill this gap, and introduced MagnifierBench, which could test the ability of MLLMs' to discern the details of small objects in high-resolution input images.
% Considering that most open-source MLLMs adopted the Pretrained Contrastive Language-Image PreTraining (CLIP) model as the visual encoder, Tong et al.\cite{tong2024MMVP} proposed the MMVP benchmark consisting of CLIP-blind pairs, finding that all tested models struggled with simple questions on visual details.
% In view of that existing benchmarks were of insufficient size and didn't cover crucial visual elements such as depth and spatial awareness, CV-Bench\cite{tong2024CV-Bench} was created to evaluate the fundamental 2D and 3D visual understanding of MLLMs using significantly more examples.
% Besides, P2GB\cite{chen2024P2GB} was designed to assess the fine-grained visual capabilities especially under text-rich scenarios, requiring a comprehensive ability in recognition and image text content understanding. 
% Moreover, VisualCoT \cite{shao2024VisualCoT} was constructed with visual chain-of-thought prompts, allowing MLLMs to focus on specific regions within a complete image, similar to how humans comprehend intricate visual information.

\textbf{Nuanced Vision-language Alignment}:Nuanced vision-language alignment involves interpreting complex interactions between visual and textual information, grasping subtle meanings, and aligning semantics between images and text. 
Winoground \cite{thrush2022Winoground} was designed to require models to match two images with two captions that contained the same set of words in different orders. However, the scale of Winoground was restricted by its costly curation, and it lacked a focus on linguistic phenomena.
To address these limitations, VALSE \cite{VALSE} and VLChecklist \cite{zhao2023VL-CheckList} examined how MLLMs understood visual-linguistic concepts by converting real captions into confusing alternatives.
By modifying textual representations related to relationships, attributes, and order, ARO \cite{yuksekgonul2023ARO} assessed whether MLLMs could achieve fine-grained visual-language alignment on key concepts.
% By modifying textual representations related to relationships, attributes, and order, ARO\cite{yuksekgonul2023ARO} assessed MLLMs' understanding of fine-grained vision-language alignment in .
Moreover, Eqben\cite{wang2023Eqben} assessed whether MLLMs were sensitive to visual semantic changes by making minimal semantic changes in images, but the image diversity was limited by virtual engines. 
To produce diverse images that met the requirements, Peng et al.\cite{peng2024SPEC} developed a benchmark called SPEC, utilizing a progressive pipeline to synthesize images that varied in a specific attribute while ensuring consistency in other aspects.
% Winoground \cite{thrush2022Winoground} was designed to require MLLMs to match two images with two captions that contained the same set of words in different orders. However, scale of Winoground was restricted by its costly curation and it lacked focus on linguistic phenomena.
% To address the above limitations, VALSE \cite{VALSE} and VLChecklist \cite{zhao2023VL-CheckList} examined how MLLMs understand visual-linguistic concepts by converting real captions into confusing alternatives.
% % To address the above limitations, VALSE\cite{VALSE} and VLChecklist\cite{zhao2023VL-CheckList} investigated visual-linguistic concepts Understanding in MLLMs by transforming real captions into confusing alternatives. 
% By modifying textual representations related with relationships, attributes, and order, ARO assessed MLLMs' understanding of fine-grained visual-linguistic concepts.
% % Considering it was unclear how well MLLM encode the compositional relationships between objects and attributes, ARO\cite{yuksekgonul2023ARO} was proposed to evaluate MLLM’s fine-grained linguistic concepts understanding. 
% Eqben assessed whether MLLM was sensitive to visual semantic changes by making visual minimal semantic changes, but the image diversity was limited by virtual engines. To produce diverse images that meet the requirements, Peng et al. \cite{peng2024SPEC} developed a benchmark called SPEC, utilizing a progressive pipeline to synthesize images that vary in a specific attribute while ensuring consistency in other aspects.

\subsection{Image Understanding}
The image understanding task involves analyzing visual content to extract meaningful information, which includes grasping the context of scenes and integrating visual details with textual information to generate coherent descriptions and insights.

\textbf{Multi-image Understanding}:
Multi-image understanding requires MLLMs to compare, analyze, and comprehend the relationships or variations among multiple images, thereby enabling more comprehensive insights into visual content.
Mementos\cite{wang2024Mementos} was designed to assess MLLMs' sequential image understanding abilities, but it primarily focused on state changes among images, while neglecting other aspects.
To address this issue, MileBench \cite{song2024MileBench} and MuirBench \cite{wang2024MuirBench} were constructed to assess multi-image cognition through a variety of task types.
% To address this issue, MileBench\cite{song2024MileBench} and MuirBench\cite{wang2024MuirBench} were constructed to provide a more thorough assessment of multi-image cognition.
However, these benchmarks fell short in terms of task depth and breadth. 
Furthermore, MMIU \cite{meng2024MMIU} provided a comprehensive evaluation by including a large number of test samples that spanned a wide range of multi-image tasks and relationships.
% Furthermore, MMIU\cite{meng2024MMIU} offered a comprehensive evaluation by encompassing massive test samples spanning various multi-image tasks and image relationships. 
Besides, Kil et al.\cite{kil2024CompBench} focused on tasks involving relativity and comparison between multiple visual input, then introduced COMPBENCH to evaluate the comparative understanding capabilities of MLLMs. COMPBENCH required MLLMs to answer questions based on a pair of visually or semantically related images.
% Besides, Kil et al.\cite{kil2024CompBench} paid attention to tasks that involve relativity and comparison between multiple visual inputs, and introduced COMPBENCH to evaluate the comparative reasoning capabilities of MLLMs.
% CompBench required MLLMs to answer questions according to a pair of visually or semantically relevant images

\textbf{Implication Understanding:} Understanding the meaning of images requires not only intuitive observation but also an exploration of the human emotions and cultural contexts they reflect.
II-Bench\cite{liu2024II-Bench} and ImplicitAVE\cite{zou2024ImplicitAVE} aimed to assess MLLMs’ higher-order perceptual abilities with visual implications. To evaluate the emotional perception capabilities of MLLMs, FABA-Bench \cite{li2024FABA-Bench} was designed for facial affective behavior analysis. FABA-Bench required MLLM to recognize facial expressions and movements, which are critical to understanding an individual’s emotional states and intentions.

\textbf{Image Quality and Aesthetics Perception}:Image quality and aesthetics perception involves assessing image quality, perceiving visual distortions, and understanding low-level attributes such as color, lighting, composition, and style. It also relates to the aesthetics and design sense of photographs. Q-Bench\cite{wu2024Q-Bench} explored the potential of MLLMs in low-level perception abilities. 
To highlight subtle differences or similarities that might not be evident when images were viewed in isolation, Q-Bench+\cite{zhang2024Q-Bench+} extended the evaluation of low-level perception from single images to image pairs. 
To better align with human aesthetics, comprehensive aesthetic evaluation benchmarks AesBench\cite{huang2024AesBench} and UNIAA\cite{zhou2024UNIAA} were constructed to systematically evaluate the aesthetic abilities of MLLMs. Besides, Lin et al.\cite{lin2024DesignProbe} proposed DesignProbe to comprehensively assess design capabilities of MLLMs from both the element level and the overall design level.

\section{Cognition and Reasoning}
\label{sec:cr}

MLLMs' cognitive and reasoning abilities encompass the model's capacity for advanced processing and complex inference beyond basic perception and understanding. Cognitive abilities involve integrating and manipulating extracted information to form coherent representations, while reasoning abilities focus on drawing logical conclusions and solving problems. Strong cognitive and reasoning abilities enable MLLMs to perform effective logical inference in complex tasks.
\subsection{General Reasoning}
The reasoning ability of MLLMs involves extracting and inferring relevant information from visual and textual inputs to draw logical conclusions and answer questions. This section introduces benchmarks for evaluating MLLMs' general reasoning capabilities, focusing on three key areas: visual relation reasoning, vision-indispensable reasoning, and context-related reasoning. 

\textbf{Visual Relation:}
Evaluating MLLMs' reasoning about visual relations primarily involves spatial and causal relationships. Regarding spatial relationships, these benchmarks mainly focus on evaluating MLLMs' understanding of the spatial arrangement, relative positions, and interactions of objects~\cite{jin2024llava-vsd}.
% Evaluating MLLMs' reasoning of visual spatial relationships~\cite{jin2024llava-vsd} involves assessing their abilities to understand the spatial arrangement, relative positions, and interactions of objects.
VSR\cite{VSR} and What's Up\cite{kamath2023what'sup} were developed to test MLLMs' ability to reason about spatial relations in natural image-text pairs. However, these benchmarks primarily focused on classification tasks rather than comprehending relations within scenes. To assess more complex and general relational reasoning of MLLMs, Wang et al.\cite{wang2024CRPE} introduced the Circular-based Relation Probing Evaluation (CRPE), the first benchmark to encompass all elements of relation triplets (subject, predicate, object). Recognizing that interactions and associations between distinct objects remained a significant challenge for MLLMs, Nie et al.\cite{nie2024MMRel} developed Multi-Modal Relation Understanding (MMRel), a large-scale benchmark consisting of data with well-defined inter-object relations. Additionally, GSR-BENCH\cite{rajabi2024GSR-BENCH} extended the What's Up benchmark by incorporating bounding box annotations and depth information to better evaluate grounded spatial relation understanding. Similarly, Chen et al.\cite{cheng2024SpatialRGPT} proposed SpatialRGBT-Bench, a benchmark that incorporated ground-truth 3D annotations and flexibly integrated depth information. In terms of causal relationships, causal reasoning refers to identifying the relationship between a cause and its effect. To explore MLLMs' ability to perform multimodal causal reasoning tasks, Li et al.\cite{li2024MuCR} proposed a benchmark called MuCR. This benchmark challenged MLLMs to infer semantic cause-and-effect relationships based solely on visual cues such as action, appearance, clothing, and environment.

\textbf{Context-related Reasoning:} 
Context-related visual comprehension requires MLLMs to effectively leverage contextual knowledge in solving visual problems. 
To assess the ability of MLLMs to answer context-dependent questions, CODIS \cite{luo2024CODIS} was created, requiring MLLMs to use context from free-form text to enhance visual comprehension.
% To assess the ability o MLLMs to answer context-dependent questions, CODIS\cite{luo2024CODIS} was created to evaluate MLLMs' ability to use context from free-form text to enhance visual comprehension. 
Recognizing that existing MLLMs often trusted what they saw but struggled to understand presuppositions in sentences, Li et al.\cite{li2024CFMM} introduced CFMM, a counterfactual reasoning benchmark designed to assess MLLMs' ability to make presuppositions based on established facts. 
Zong et al. \cite{zong2024VL-ICLBench} developed VL-ICL Bench, a benchmark suite designed to specifically evaluate VLLMs' in-context learning, which involved utilizing contextual information to complete new tasks.
% Zong et al.\cite{zong2024VL-ICLBench} developed VL-ICL Bench, a benchmark suite designed to specifically evaluate VLLMs' in-context learning, which is the ability to quickly adapt to new tasks using few-shot examples as prompts.
% Context-dependent visual comprehension requires MLLMs to effectively leverage contextual knowledge in solving visual problems. CODIS was created to evaluate MLLMs' ability to use context from free-form text to enhance visual comprehension. Recognizing that existing MLLMs often trust what they see but struggle to understand presuppositions in sentences, Li et al.\cite{li2024CFMM} introduced CFMM, a counterfactual reasoning benchmark designed to assess MLLMs' ability to make presuppositions based on established facts. Additionally, Zong et al.\cite{zong2024VL-ICLBench} developed VL-ICL Bench, a benchmark suite specifically designed to evaluate VLLMs' in-context learning—the ability to quickly adapt to new tasks using few-shot examples as prompts.

\textbf{Chain-of-Thought Reasoning:}Multi-modal Chain-of-Thought (CoT) requires models to leverage knowledge from both textual and visual modalities for step-by-step reasoning. To diagnose the multi-hop reasoning ability and interpretability of multi-modal large language models (MLLMs), Lu et al.\cite{lu2022ScienceQA} generated lectures and explanations as CoT to mimic the multi-hop reasoning process for answering questions and presented Science Question Answering (SCIENCEQA). SCIENCEQA consisted of questions covering diverse science topics, along with annotations of their answers and corresponding lectures and explanations, demonstrating the utility of CoT in improving the question-answering performance of MLLMs. Besides, Shao et al.\cite{shao2024VisualCoT} introduced the visual chain-of-thought benchmark for evaluating MLLMs in scenarios where they need to focus on specific local regions or reasons to identify objects. However, these benchmarks were relatively simple and focused only on certain specific domains. Therefore, Chen et al.\cite{chen2024M3CoT} introduced M3CoT to address these limits, advancing the multi-domain, multi-step, and multi-modal CoT.

\textbf{Vision-Indispensable Reasoning:}
Recognizing that MLLMs may rely on language priors rather than visual information when answering questions, some works have aimed to compel MLLMs to prioritize visual data. 
Goyal et al.\cite{goyal2017VQAv2} introduced VQAv2, which consisted of pairs of similar images that led to different answers. However, this approach did not effectively handle open-ended questions. 
In response, CLEVR\cite{johnson2016CLEVR} was designed with open-ended question answering. It also ensured that external information sources, such as commonsense knowledge, did not influence answer accuracy. Nonetheless, CLEVR's reliance on synthetic images overlooked the realism and diversity found in natural photographs. To address these limitations, GQA\cite{hudson2019GQA} was developed, offering well-defined semantic representations along with the rich semantic and visual complexity of real-world images. Additionally, Chen et al.\cite{chen2024MMStar} introduced MMStar, a vision-indispensable benchmark that covered a wide range of tasks and difficulty levels.
% Some works have aimed to compel MLLMs to prioritize visual information over relying on language priors. Goyal et al.\cite{goyal2017VQAv2} introduced VQAv2, which features pairs of similar images that lead to different answers. However, this approach failed to address open-ended questions, resulting in an unbalanced answer distribution. In contrast, CLEVR\cite{johnson2016CLEVR} was designed to ensure that external information sources, like commonsense knowledge, do not improve the accuracy of answers. Yet, CLEVR's reliance on synthetic images overlooked the realism and diversity found in natural photographs. To overcome these limitations, GQA\cite{hudson2019GQA} was developed, offering both well-defined semantic representations and the rich semantic and visual complexity of real-world images. Additionally, Chen et al. created MMStar\cite{chen2024MMStar}, a vision-indispensable multi-modal benchmark that spans diverse tasks and levels of difficulty.

\subsection{Knowledge-based Reasoning}
Evaluating MLLMs' ability to utilize knowledge is crucial for ensuring their effectiveness in complex tasks and enhancing their real-world performance. These benchmarks mainly focus on two key aspects. 
One aspect is knowledge-based question answering, which tests MLLMs' ability to handle questions that require structured or extensive external knowledge. The other aspect is knowledge editing, which assesses MLLMs' accuracy and consistency in updating and maintaining knowledge content.
% The first aspect is knowledge-based question answering, which tests MLLMs' ability to handle questions that require structured or extensive external knowledge.
% % The first aspect is knowledge-based question answering, which tests MLLMs on their ability to handle questions requiring structured or extensive external knowledge. 
% The second aspect is knowledge editing, which assesses the MLLMs' accuracy and consistency in updating and maintaining knowledge content.

\textbf{Knowledge-based Visual Question Answering:}
Knowledge-based visual question answering requires various types of knowledge, including explicit fact-based information from knowledge bases, commonsense knowledge about human behavior, and general external knowledge.
The earliest benchmarks in this area were KB-VQA\cite{wang2015KBVQA} and FVQA\cite{wang2017FVQA}, but they were limited by their use of "closed" knowledge sources. OK-VQA\cite{marino2019OK-VQA} advanced these benchmarks by providing a larger scale and higher quality of questions and images, utilizing "open domain" knowledge rather than a fixed source. However, OK-VQA still relied on simple lookup knowledge and required minimal reasoning. To overcome this limitation, A-OKVQA\cite{schwenk2022A-OKVQA} was introduced, incorporating more commonsense knowledge and demanding greater reasoning. Furthermore, Wang et al.\cite{wang2024SOK-Bench} developed SOK-Bench to evaluate MLLMs' situated and open-world commonsense reasoning in videos.
% Knowledge-based visual question answering (VQA) requires various forms of external knowledge, including explicit fact-based information from knowledge bases, commonsense knowledge about human behavior, understanding of physical principles, and visual knowledge. The earliest knowledge-based VQA benchmarks were KB-VQA\cite{wang2015KBVQA} and FVQA\cite{wang2017FVQA}, but the knowledge they used was "closed" and limited. OK-VQA improved on prior benchmarks by offering a larger scale and higher quality questions and images, and it utilized "open domain" knowledge rather than knowledge from a fixed source. Despite this, OK-VQA still relied on simple lookup knowledge and required minimal reasoning. To address these limitations, A-OKVQA\cite{schwenk2022A-OKVQA} was developed to incorporate more commonsense knowledge and require greater reasoning. Additionally, Wang et al.\cite{wang2024SOK-Bench} introduced the benchmark SOK-Bench for evaluating MLLMs' situated and open-world commonsense reasoning in videos.

\textbf{Knowledge Editing:}
Knowledge editing refers to the ability to update outdated, unknown, or incorrect information within MLLMs. 
The benchmark MMEdit, proposed by Cheng et al.\cite{cheng2024MMEdit}, provided a platform for testing the editability of MLLMs. However, it primarily focused on coarse-grained knowledge, which often failed to accurately represent fine-grained entities and scenarios in the real world. 
To address this limitation, Cheng\cite{cheng2024MMEdit} introduced MIKE, a comprehensive and challenging benchmark for fine-grained multimodal entity knowledge editing. Meanwhile, VLKEB \cite{huang2024VLKEB} expanded the evaluation of knowledge editing portability, demonstrating MLLMs' ability to effectively apply edited knowledge in related contexts. 
Despite that, these benchmarks overlooked the organization of multimodal knowledge and lacked a precise definition of multimodal knowledge editing. To fill this gap, MC-MKE\cite{MC-MKE} was developed as a benchmark to evaluate the reliability, locality, generality, and consistency of MLLMs across different editing formats.
% Knowledge editing refers to the ability to update outdated, unknown, or incorrect knowledge within MLLMs. The benchmark MMEdit proposed by Cheng et al.\cite{cheng2024MMEdit} provided a platform to test MLLMs' editability. However, it primarily focused on coarse-grained knowledge, which often fails to accurately represent fine-grained entities and scenarios in the real world. To address this limitation, Cheng\cite{cheng2024MMEdit} introduced MIKE, a comprehensive and challenging benchmark for fine-grained multimodal entity knowledge editing. Meanwhile, VLKEB\cite{huang2024VLKEB} extended portability evaluation, demonstrating MLLMs' ability to effectively apply edited knowledge in related contexts. However, it overlooked the organization of multimodal knowledge and lacked a precise definition of multimodal knowledge editing. To fill these gaps, MC-MKE\cite{MC-MKE} was developed, a benchmark that evaluates the reliability, locality, generality, and consistency of MLLMs across different editing formats.

\subsection{Intelligence\&Cognition:}
Inspired by the development of human intelligence, some benchmarks leverage cognitive and educational theories to assess the intelligence of MLLMs. For instance, intelligence tests featuring abstraction visual reasoning and various levels of mathematical problems are used to evaluate MLLMs' logical reasoning capabilities. Additionally, multidisciplinary questions from various educational periods are employed to assess MLLMs' ability to integrate diverse knowledge and apply complex reasoning skills to solve intricate problems. These approaches are crucial for understanding and enhancing the cognitive and problem-solving capabilities of MLLMs.

\textbf{Intelligent Question Answering:}
Intelligent question answering aims to explore the intelligence of MLLMs through cognitive science perspectives.
One key aspect is abstract visual reasoning (AVR)—the ability to discern relationships among patterns in images and predict subsequent patterns.
% One key aspect examined is abstract visual reasoning (AVR)—the ability to discern relationships among patterns in images and extrapolate to predict subsequent patterns. 
RAVEN\cite{zhang2019RAVEN} tested abstract visual reasoning primarily with mathematical patterns over predefined geometric shapes, but its evaluation scope was limited as it addressed only a single task type.
% RAVEN tested abstract visual reasoning primarily with mathematical patterns over predefined geometric input shapes, but its evaluation scope was limited due to addressing only a single task type. 
Therefore, MARVEL\cite{jiang2024MARVEL} and VCog-Bench\cite{cao2024VCog-Bench} were introduced to evaluate MLLMs across multi-dimensional AVR tasks, but remained confined to AVR, neglecting other dimensions of cognition. 
To integrate cognitive science principles for a comprehensive understanding of MLLMs’ intelligence, Song\cite{song2024M3GIA} identified five key cognitive factors based on the well-recognized Cattell-Horn-Carroll (CHC) model and introduced M3GIA, the first comprehensive cognitive-driven benchmark designed to evaluate the general intelligence of MLLMs.

\textbf{Mathematical Question Answering:}
Using mathematics problems to evaluate MLLMs is essential for assessing their logical reasoning capabilities, as these problems require complex reasoning, pattern recognition, and abstract thinking. Such tasks help determine if MLLMs can apply rules, discover patterns, and perform sophisticated reasoning.
Geometry3K\cite{lu2021Geometry3K} was introduced to evaluate capabilities in solving geometry problems, but it had a narrow focus on specific aspects of plane geometry.
% Geometry3K\cite{lu2021Geometry3K} was introduced to evaluate capabilities in solving geometry problems, but it focused narrowly on specific aspects of plane geometry. 
% % This narrow focus limited the assessment of broader mathematical abilities. 
To address this issue, Lu et al.\cite{lu2024MathVista} collected multiple datasets to construct an integrated benchmark, MathVista, which covered a range of mathematical tasks, such as functions and solid geometry. However, MathVista lacked a detailed classification of mathematical subdomains and emphasized visual abilities more than pure mathematical reasoning. 
Consequently, Math-V\cite{wang2024Math-Vision} and MathVerse\cite{MathVerse} were developed, confining data to specific mathematical subjects and focusing on mathematical reasoning abilities. 
Additionally, Fan et al.\cite{fan2024NPHardEval4V} proposed NPHardEval4V, a benchmark that used algorithmic problems and converted their textual descriptions into visual representations, aimed at evaluating the pure reasoning capabilities of MLLMs. 
Motivated by the idea that "if a model truly understands a problem, it should perform robustly across various tasks related to that problem," Zhou et al.\cite{zhou2024MATHCHECK} introduced MATHCHECK-GEO, a benchmark focused on the universality of tasks and the robustness of question formulation within visual contexts. It was designed to test whether a model's performance is consistent across different tasks related to the same problem.
% Geometry3K\cite{lu2021Geometry3K} was introduced to evaluate MLLMs' geometry problem solving, but focused narrowly on specific aspects of plane geometry. This limited the evaluation of broader mathematical capabilities, e.g., functions and solid geometry.
% To integrate various math-related visual reasoning tasks, Lu et al.\cite{lu2024MathVista} collected multiple datasets to construct an integrated benchmark MathVista, designed to evaluate mathematical reasoning in visual contexts. 
% However, it lacked a detailed classification of mathematical subdomains and also emphasized other visual abilities rather than mathematical reasoning.
% Therefore, Math-V\cite{wang2024Math-Vision} and MathVerse\cite{MathVerse} were constructed with data confined to a narrow scope of math subjects.
% Besides, Fan et al.\cite{fan2024NPHardEval4V} proposed NPHardEval4V by retaining algorithmic problems and converted their textual descriptions into visual representations, aimed at addressing the existing gaps in evaluating the pure reasoning abilities of MLLMs.
% Motivated by \emph{ if a model really understands a problem, it should robustly work across various tasks about this problem},
% Zhou et al.\cite{zhou2024MATHCHECK} introduced MATHCHECK, a benchmark focused on the universality of tasks and robustness of question formulation within visual contexts. 

\textbf{Multidisciplinary Question Answering:} 
Evaluating MLLMs using multidisciplinary questions from various educational stages assesses their ability to integrate and apply knowledge across different domains. This approach tests MLLMs' reasoning and problem-solving skills in diverse contexts, providing a comprehensive measure of general intelligence and cognition.
ScienceQA\cite{lu2022ScienceQA} was a benchmark containing multimodal science questions with rich domain diversity. While it covered a range of disciplines, most of the questions were at the elementary to middle school level, which limited its depth. 
To address this issue, M3Exam\cite{zhang2023M3Exam} was proposed with a multilevel structure, featuring exams from three critical educational stages to comprehensively assess MLLMs' proficiency at different levels. 
Additionally, SceMQA\cite{liang2024SceMQA} focused on college entrance-level problems. Due to the importance of this stage, SceMQA comprised questions with answers accompanied by more detailed explanations.
Intended to evaluate expert-level understanding, Yue et al.\cite{yue2024MMMU} introduced MMMU, which included problems from college exams, quizzes, and textbooks across six common disciplines. 
However, these benchmarks were primarily available in English, restricting the evaluation to a single language. Therefore, CMMMU\cite{zhang2024CMMMU}, CMMU\cite{he2024CMMU}, and MULTI\cite{zhu2024MULTI} were created to evaluate multi-discipline and multi-type question understanding and reasoning in Chinese.
% ScienceQA\cite{lu2022ScienceQA} was a benchmark containing multimodal science questions with rich domain diversity. While it covered diverse disciplines, the majority of the questions were at the elementary to the middle school level, thus falling short in depth. 
% M3Exam\cite{zhang2023M3Exam} was proposed with multilevel structure, featuring exams from three critical educational periods to comprehensively assess a MLLMs' proficiency at different levels.
% Additionally, SceMQA \cite{liang2024SceMQA} focused on college entrance-level problems, with a high proportion accompanied by detailed explanations.
% Aimed to assess the expert-level understanding capability of MLLMs, Yue et al.\cite{yue2024MMMU} introduced MMMU, which contained problems sourced from college exams, quizzes, and textbooks spanning six common disciplines.
% However, these benchmarks were predominantly available in English, which imposed limitations on the comprehensiveness of the evaluation. To address this issue, CMMMU\cite{zhang2024CMMMU}, CMMU\cite{he2024CMMU} and MULTI\cite{zhu2024MULTI} were created for multi-discipline and multi-type question understanding and reasoning in Chinese. 

\section{Specific Domains}
\label{sec:sd}

This section focuses on MLLMs' capabilities in specific tasks and applications, such as their ability to integrate complex visual and textual information, adapt to decision-making roles in dynamic environments, and effectively process diverse cultural and linguistic data. It then extends to discuss the practical applications of MLLMs, highlighting their impact on various sectors such as medicine, industry, and autonomous driving. By providing an overview of these benchmarks, this section aims to underscore the advancements in evaluating MLLMs' performance and their potential to address real-world challenges across different domains.
% This chapter introduces benchmarks that explore MLLMs' capabilities in specific domains, focusing on their performance in text-rich visual question answering, adaptability as decision-making agents, understanding and processing of diverse cultures and languages, and their application across various fields, including industry, medicine, remote sensing, and autonomous driving.

\subsection{Text-rich VQA}
Evaluating MLLMs in text-rich visual question answering is crucial for understanding how well models interpret and integrate textual and visual information within images. This evaluation covers several aspects, including the accuracy of text recognition, contextual understanding, and the ability to synthesize information from both modalities. In addition to text comprehension, it also requires an understanding of layout and structure to effectively analyze multimodal documents, charts, and HTML.
% Evaluating MLLMs in Text-rich VQA is crucial for understanding how well models interpret and integrate textual and visual information within images. This includes aspects like accuracy in text recognition, contextual understanding, and the ability to synthesize information from both modalities. The evaluation also looks at the model's capability to comprehend and analyze multimodal documents, charts, and HTML, requiring effective integration of text and visuals.

\textbf{Text-oriented Question Answering:}
Some works evaluated MLLMs' effectiveness in text-related visual tasks, such as text recognition and scene text-centric visual question answering. 
Singh et al.\cite{singh2019TextVQA} introduced TextVQA, which contained questions requiring the model to read and reason about the text in the image to provide answers. However, the short answers provided in TextVQA were insufficient for a comprehensive description of the image.
To address this issue, TextCaps\cite{sidorov2020TextCaps} extended the length of sentences in the answers and involved many switches between OCR and vocabulary tokens.
Despite these efforts, existing benchmarks could be time-consuming, and inaccurate annotations in some datasets made accuracy-based evaluation less precise. In response, Liu et al.\cite{liu2024OCRBench} developed OCRBench to facilitate the accurate and convenient evaluation of MLLMs’ OCR capabilities. 
To quantitatively assess visual reasoning capabilities in text-rich and high-resolution scenarios, Chen et al.\cite{chen2024P2GB} constructed a challenging benchmark, P2GB. This benchmark included comprehensive image understanding, fine-grained recognition, and image text content understanding.
% To quantitatively assess visual reasoning capabilities in text-rich and high-resolution scenarios, Chen et al.\cite{chen2024P2GB} constructed a challenging benchmark, P2GB, which included comprehensive image understanding with fine-grained recognition and image text content understanding. 
In order to cover a broader spectrum of text-rich scenarios, SEED-Bench-2-Plus\cite{li2024SEED-Bench-2-Plus} was developed to evaluate MLLMs’ performance in comprehending text-rich visual data across a wide range of real-world scenarios.
% Some works evaluate MLLMs' effectiveness in text-related visual tasks, such as text Recognition and scene text-centric VQA. Singh et al.\cite{singh2019TextVQA} introduced TextVQA containing questions which require the model to read and reason about the text in the image to be answered, but short answers in TextVQA were insufficient to describe the image comprehensively. In contrast to TextVQA, TextCaps\cite{sidorov2020TextCaps} required generating long sentences and involves many switches between OCR and vocabulary tokens. However, existing benchmarks can be time-consuming, and the inaccurate annotations in some datasets made accuracy based evaluation less precise. Realizing these limitations, Liu et al.\cite{liu2024OCRBench} developed OCRBench to facilitate the accurate and
% convenient evaluation of MLMMs’ OCR capabilities. To quantitatively assess the visual reasoning capabilities under text-rich or high-resolution scenarios, Chen et al.\cite{chen2024P2GB} constructed a challenging benchmark P2GB, which included comprehensive image understanding with fine-grained recognition and image text content understanding. To cover a wider spectrum range of text-rich scenarios, SEED-Bench-2-Plus\cite{li2024SEED-Bench-2-Plus} was constructed to focus on evaluating MLLMs’ performance in comprehending text-rich visual data, which covered a wide spectrum range of text-rich scenarios in the real world.

\textbf{Document-oriented Question Answering:} Document-oriented question answering requires MLLMs not only to read text but also to interpret it within the layout and structure of the document. 
Mathew et al.\cite{mathew2021infographicvqa} introduced InfographicVQA, which included questions requiring the combination of multiple cues. 
Although InfographicVQA showcased significant diversity in topics and designs, it still preferred using visual aids over long text passages.
To address this limitation, Single Page DocVQA (SPDocVQA)\cite{mathew2021SPDocVQA} was introduced as a more diverse benchmark, featuring documents of various types and origins created over several decades. However, it was built exclusively on single-page document excerpts and was limited to several domains represented in the Industry Documents Library. 
MP-DocVQA\cite{tito2023MPDocVQA} extended this by including preceding and following pages of documents, but the additional pages often served as mere distractors.
To overcome these issues and establish a more practical and enduring benchmark, DUDE\cite{vanlandeghem2023DUDE} was created as a large-scale, multi-page, multidomain, multi-industry benchmark for evaluating MLLMs in document understanding. 
Additionally, to evaluate the ability of MLLMs to understand long multimodal documents, MM-NIAH\cite{wang2024MM-NIAH} was designed to systematically assess long context capability of MLLMs using the Needle-In-A-Haystack (NIAH) test.
% Addressing these issues, the most diverse benchmark to date is Single Page DocVQA (SPDocVQA), which contained mixed-origin documents of different types created over several decades. However, it was built exclusively on single-page document excerpts and was limited to several domains represented in the Industry Documents Library. MP-DocVQA extended this by including preceding and following pages of documents; however, the additional pages often served as mere distractors. To address these issues and establish a more practical and enduring benchmark, DUDE\cite{vanlandeghem2023DUDE} was created as a novel, large-scale, multi-paged, multidomain, multi-industry DocVQA benchmark for evaluating MLLMs in document understanding. Additionally, to evaluate the ability to understand long multimodal documents, MM-NIAH\cite{wang2024MM-NIAH} was designed to systematically assess the comprehension capability of MLLMs through the Needle-In-A-Haystack (NIAH) test.
% DocVQA\cite{mathew2021DocVQA} comprised document images of industry/business documents, and questions that require understanding document elements such as text passages, forms, and tables. Considering infographics in doucuments are unique in their combined use, and purposeful arrangement of visual and textual elements, 

\textbf{Chart-oriented Question Answering}:
Chart understanding plays a pivotal role when applying MLLMs to real-world tasks such as analyzing scientific papers or financial reports. 
ChartQA\cite{masry2022ChartQA} was a large-scale benchmark involving visual and logical reasoning over charts, but it had limited chart types. 
To scale up, Li et al.\cite{li2023SciGraphQA} utilized Arxiv papers and constructed SciGraphQA with multi-turn question-answer conversations on scientific graphs. MMC\cite{liu2024MMC-Benchmark} and ChartBench\cite{xu2024ChartBench} further expanded the types and tasks of chart data. However, the question design in these benchmarks remained relatively simplistic. To further validate MLLMs' capabilities in more complex reasoning tasks involving chart data, Xia et al.\cite{xia2024ChartX} proposed the benchmark ChartX for comprehensive chart understanding and reasoning. 
However, ChartX yielded artificial charts generated by GPT-4\cite{achiam2023gpt4}, leading to a narrow distribution. Therefore, CharXiv\cite{wang2024CharXiv} was proposed for evaluating MLLMs' understanding of real-world scientific charts, including complex compositions with multiple subplots collected from arXiv. To further improve the quality, Roberts et al.\cite{roberts2024SciFIBench} presented SciFIBench, a benchmark for scientific figure interpretation. They adopted adversarial filtering when curating negative examples for each question to increase difficulty, while also performing human verification on every question to ensure high-quality content.
Moreover, in order to assess MLLMs' understanding of charts from various perspectives, Fan et al.\cite{fan2024CHOPINLLM} proposed CHOPINLLM, including three different levels of questions (literal, inferential, and reasoning QAs).

\textbf{Html-oriented Question Answering:}
Web pages present a complex interplay of visual and textual information, along with interactive elements, requiring MLLMs to possess rigorous understanding abilities over hierarchical structures and contextual relationships. Web2Code\cite{yun2024Web2Code} was a benchmark with web pages based on instruction-response pairs. The responses included structured questions and answers about the webpage. However, web elements are often small, numerous, and scattered across the page, demanding fine-grained recognition and accurate spatial reasoning. To address these limitations, Liu et al.\cite{liu2024VisualWebBench} introduced VisualWebBench, which assessed MLLMs at three levels: website-level, element-level, and action-level. Additionally, following the principle "What I cannot create, I do not understand," Plot2Code\cite{wu2024Plot2Code} evaluated MLLMs' ability to generate code that effectively rendered a provided image from HTML files, further showcasing their multimodal understanding and reasoning capabilities.
% Web pages present a complex interplay of visual and
% textual information, along with interactive elements, requiring models to possess rigorous
% understanding abilities over hierarchical structures and contextual relationships. Web2Code\cite{yun2024Web2Code} was a benchmark with  webpages based instruction-response pairs. The responses consist of not only the HTML code, but also the structured questions and answers about the webpage. However, web elements are often small, numerous, and scattered across the page, demanding fine-grained recognition and accurate spatial reasoning and grounding. To address these limitations, Liu et al.\cite{liu2024VisualWebBench} introduced VisualWebBench, which assessed MLLMs at three levels: website-level, element-level, and action level. Additionally, in view of "What I cannot create, I do not understand", Plot2Code\cite{wu2024Plot2Code} evaluated the capability of MLLMs to generate code that rendered a provided image from HTML files effectively, which can further showcase their multi-modal understanding and reasoning powers.

\subsection{Decision-making Agents}
Decision-making agents expect MLLMs to possess human-level planning and scheduling abilities, which are fundamental for making informed decisions and taking appropriate actions in complex environments. This capability holds significant potential for addressing real-world problems.
% Decision-making agent expects MLLMs to have ability of human-level planning, a fundamental ability for making informed decisions in complex environments, and solving a wide range of real-world problems.

\textbf{Embodied Decision-making:}
Embodied Decision-making requires MLLMs to be able to integrate sensory inputs and interact with the environment in a way that mimics human physical experiences. 
OpenEQA\cite{OpenEQA2023} was the first benchmark for embodied question answering, supporting both episodic memory and active exploration use cases. However, it focused solely on the answers provided by MLLMs and did not consider the intermediate reasoning processes.
Chen et al.\cite{chen2023PCA-EVAL} argued that it was essential to enable multi-dimensional evaluation of the decision-making process, encompassing perception, reasoning, and action perspectives, rather than relying solely on final rewards or success rates. They proposed PCA-EVAL for evaluating the embodied decision-making ability of MLLMs from different perspectives, including three complex scenarios: autonomous driving, domestic robotics, and open-world games. However, these benchmarks were limited by the small number of handcrafted questions and reliance on single-image visual observation.
% However, these benchmarks were constrained by a limited number of handcrafted questions and reliance on single-image visual observation. 
To study the embodied planning and decision-making capabilities of MLLMs more systematically, Chen et al.\cite{chen2024EgoPlan-Bench} used large-scale egocentric videos reflecting daily human activities from a first-person perspective to construct EgoPlan-Bench.
% Chen et al.\cite{chen2024EgoPlan-Bench} leveraged large-scale egocentric videos that authentically reflected daily human activities from the first-person perspective and constructed EgoPlan-Bench. 
EgoPlan-Bench aimed to assess MLLMs' human-level planning capabilities in real-world scenarios, featuring realistic tasks, diverse actions, and complex visual observations.
% EgoPlan-Bench aimed to evaluate MLLMs' capabilities for human-level planning in real-world scenarios, featuring realistic tasks, diverse actions, and intricate visual observations. 
Despite that, these benchmarks failed to sufficiently challenge or showcase the full potential of MLLMs in complex environments. 
To address this gap, Liu et al.\cite{liu2024VisualAgentBench} introduced VisualAgentBench (VAB), a comprehensive and pioneering benchmark specifically designed to train and evaluate MLLMs as visual foundation agents across diverse scenarios, with tasks formulated to probe the depth of MLLMs' understanding and interaction capabilities.

\textbf{Mobile Agency:}
Mobile-Agent leverages visual perception tools to accurately identify and locate both visual and textual elements within the app’s front-end interface. Drawing on the visual information, it autonomously plans and decomposes complex tasks, navigating the mobile app through each step of the operation.
% Mobile-Agent leverages visual perception tools to accurately identify and locate both the visual and textual elements within the app’s front-end interface. Based on the perceived vision context, it then autonomously plans and decomposes the complex operation task and navigates the mobile apps through operations step by step. 
To comprehensively assess Mobile-Agent’s capabilities, Wang et al.\cite{wang2024Mobile-Eval} introduced Mobile-Eval, a benchmark centered around current mainstream mobile apps. Mobile-Eval included instructions for various difficulty levels, but it primarily assessed whether MLLMs could complete the instructions, neglecting the fine-grained perception of the web interface. To fill this gap, You et al.\cite{you2024Ferret-UI} proposed Ferret-UI, which required MLLMs to explain the functionality and provide fine-grained descriptions of UI elements. However, existing benchmarks for MLM agents in interactive environments were constrained by single environments and lacked detailed and generalized evaluation methods. To overcome these limitations, Xu et al.\cite{xu2024CRAB} introduced CRAB, the first agent benchmark framework designed to support cross-environment tasks.
% Mobile-Agent leverages visual perception tools to accurately identify and locate both the visual and textual elements within the app’s front-end interface. Based on the perceived vision context, it then autonomously plans and decomposes the complex operation task, and navigates the mobile Apps through operations step by step. In order to comprehensively assess
% Mobile-Agent’s capabilities, Wang et al.\cite{wang2024Mobile-Eval} introduced Mobile-Eval, a benchmark centered around current mainstream mobile Apps. Mobile-Eval includes instructions for various difficulty levels, but it primarily assesses whether MLLMs can complete the instruction, neglecting fine-grained perception of the web interface. To fill this gap, You et al.\cite{you2024Ferret-UI} proposed Ferret-UI, requiring MLLMs to explane the functionality and fine-grained description of UI elements. However, Existing benchmarks for MLM agents in interactive environments are constrained by single environments and lack detailed and generalized evaluation methods. To overcome these limitations, Xu et al.\cite{xu2024CRAB} introduced Crab, the first agent benchmark framework designed to support cross-environment tasks.

\subsection{Diverse Cultures and Languages}
Most benchmarks primarily use English, leading to the neglect of other languages and cultures. To address this limitation, some benchmarks have been introduced to supplement data in a broader range of languages.
% Most benchmarks primarily used English, which led to the neglect of other languages. 
% To extend the linguistic boundaries of MLLMs, some works focused on supplementing data in more languages. 
CMMU\cite{he2024CMMU}, CMMMU\cite{zhang2024CMMMU}, and MULTI\cite{zhu2024MULTI} were presented for multi-modal and multi-type questions in Chinese, featuring a wider variety of question types. Besides, the Henna benchmark\cite{alwajih2024Henna} was proposed to test MLLMs in Arabic culture, while the LaVy-Bench benchmark\cite{tran2024LaVy} was designed for evaluating MLLMs' understanding of Vietnamese visual language tasks. To further enrich language diversity, MTVQA\cite{tang2024MTVQA} was proposed as the first benchmark featuring high-quality human expert annotations across 9 diverse languages. However, although MTVQA extended its linguistic range, it kept images the same, resulting in a narrow cultural representation. To address this limitation, Romero\cite{romero2024CVQA} constructed CVQA, a culturally diverse multilingual visual question answering benchmark designed to cover a rich set of languages and cultures.

\subsection{Other Applications}
Some works focused on assessing MLLMs' abilities to handle highly professional and domain-specific data, such as medicine, transportation, engineering, remote sensing, and autonomous driving. These evaluations provided insights into how well MLLMs could adapt to and process specialized information in various complex fields, highlighting their potential for applications in areas requiring deep expertise and precise knowledge.

\textbf{Geography and Remote Sensing:}MLLMs can help enhance geographic information extraction and environmental monitoring by analyzing multimodal data in the fields of geography and remote sensing.
Roberts et al.\cite{roberts2024ChartingNewTerritories} constructed a benchmark to probe the key geographic and
geospatial knowledge of a suite MLLMs. However, the diverse geographical landscapes and varied objects in remote sensing imagery were not adequately considered. To bridge this gap, Muhtar et al.\cite{muhtar2024LHRS-Bench} introduced LHRS-Bench, a benchmark for thoroughly evaluating MLLMs’ abilities in remote sensing image understanding.

\textbf{Medicine:}Medical-based benchmarks primarily assess MLLMs' ability to integrate medical knowledge with visual modalities to perform accurate medical diagnoses and recommendations.
The Asclepius benchmark\cite{wang2024Asclepius} covered a range of medical specialties, aiming to comprehensively evaluate MLLMs' capabilities across various medical fields. However, it primarily focused on 2D medical images, leaving 3D images less explored. 
Therefore, Bai et al.\cite{bai2024M3D} introduced M3D-Bench, a 3D multimodal medical benchmark that enabled automatic evaluation across eight tasks.
Furthermore, Chen et al.\cite{chen2024GMAI-MMBench} developed GMAI-MMBench, the most comprehensive general medical AI benchmark to date, featuring a well-categorized data structure and multi-perceptual granularity. 
However, these benchmarks were relatively lacking in depth, such as overlooking the rich interconnected information across many heterogeneous biomedical sensors. To address this challenge, To address this challenge, Mo et al.\cite{mo2024MultiMed} presented MultiMed, a benchmark designed to evaluate the ability of MLLMs to integrate data from multiple sources while performing several tasks simultaneously.
% The Asclepius benchmark\cite{wang2024Asclepius} covered a range of medical specialties, aiming to comprehensively assess MLLMs' capabilities across various medical fields. However, it primarily focused on 2D medical images, leaving 3D images under-explored. Therefore, Bai et al.\cite{bai2024M3D} introduced a 3D multi-modal medical benchmark, M3D-Bench, which facilitated automatic evaluation across eight tasks. Furthermore, Chen et al.\cite{chen2024GMAI-MMBench} developed the GMAI-MMBench, the most comprehensive general medical AI benchmark with well-categorized data structure and multi-perceptual granularity to date.

\textbf{Industry:}
Some benchmarks have been designed to assess MLLMs' capabilities in industrial design and manufacturing applications. Doris et al. \cite{doris2024DesignQA} proposed a benchmark called DesignQA to explore MLLMs’ understanding of design based on engineering requirement documents. DesignQA integrated information from both visual and long-text inputs, emphasizing the complexity and multimodal nature of real-world engineering tasks. To evaluate MLLMs' abilities in robotic applications, the MMRo \cite{li2024MMRo} benchmark defined four fundamental capabilities required for a robot's central processing unit and was the first diagnostic benchmark specifically designed to systematically analyze the diverse failure modes of MLLMs in robotics.
% Doris et al.\cite{doris2024DesignQA} proposed a benchmark called DesignQA to explore MLLMs’ understanding of design according to an engineering requirement document. DesignQA integrated information from both visual and long-text inputs, emphasizing the complexity and multimodal nature of real-world engineering tasks.
% To evaluate MLLMs' abilities in robotic applications, the MMRo\cite{li2024MMRo} benchmark defined four fundamental abilities required for a robot's central processing unit, and was the first diagnostic benchmark specifically designed to systematically dissect and analyze the diverse failure modes of MLLMs in robotics.

\textbf{Society:} Benchmarking MLLMs in addressing social needs and related domains is essential for evaluating their performance in real-world interactions and practical scenarios.
To address the needs of blind individuals, Gurari et al. \cite{gurari2018VizWiz} proposed VizWiz, a benchmark consisting of visual questions originating from blind people. To evaluate MLLMs' abilities in understanding information from social media interactions, Jin et al. \cite{jin2024MM-Soc} introduced MM-SOC, a benchmark designed to holistically assess MLLMs’ performance on multimodal tasks derived from online social networks. Considering the critical role of transportation in modern society, Zhang et al. \cite{zhang2024TransportationGames} developed TransportationGames, a comprehensive benchmark aimed at accurately evaluating MLLMs' capabilities in performing transportation-related tasks.
% In order to address the interests of blind people, Gurari et al.\cite{gurari2018VizWiz} proposed VizWiz, which consisted of visual questions originating from blind people. To assess MLLMs' abilities of comprehending the information with interactions in social media platforms, Jin et al.\cite{jin2024MM-Soc} introduced MM-SOC, a benchmark to holistically evaluate MLLMs’ capability in tackling multimodal tasks derived from online social networks.
% Considering that transportation plays a crucial role in modern society, Zhang et al.\cite{zhang2024TransportationGames} introduced TransportationGames, an all-encompassing benchmark to accurately evaluate the capabilities of MLLMs in
% executing transportation-related tasks. 

\textbf{Autonomous Driving:}
Autonomous driving is a rapidly developing field with immense potential to improve transportation safety and efficiency through advancements in sensor technologies and computer vision. 
Qian et al.\cite{qian2024NuScenes-QA} constructed the first VQA benchmark for autonomous driving scenarios, named NuScenes-QA. It addressed visual question answering and provided richer visual information, including images and point clouds. However, NuScenes-QA primarily focused on scene-level driving tasks and could not cover all the reasoning processes involved in driving. In contrast, DriveLM-Data\cite{sima2024DriveLM-Data} was a more comprehensive benchmark, which encapsulated perception, prediction, and planning for autonomous driving.
% Autonomous driving is a rapidly developing field with immense potential to improve transportation safety and efficiency with advancements in sensor technologies and computer vision. Qian et al.\cite{qian2024NuScenes-QA} constructed the first VQA benchmark for autonomous driving scenario, named NuScenes-QA. It tackled high-level question answering and provided richer visual information, including images and point clouds. 
% Moreover, Sima et al.\cite{sima2024DriveLM-Data} constructed DriveLM-Data-DATA, a benchmark that encapsulated perception, prediction, and planning for autonomous driving.

\section{Key Capabilities}
\label{sec:kc}

These benchmarks evaluated dialogue capabilities, including handling extended dialogues and accurately following instructions, as well as assessing the model's level of hallucination and trustworthiness. Such capabilities are crucial for ensuring that MLLMs perform effectively across a range of real-world applications and can adapt to various practical scenarios.
% This section focus on important aspects related to user experience during usage. These include dialogue capabilities, specifically handling long contexts and accurately following instructions, as well as the model’s level of hallucination and trustworthiness. Such capabilities are essential for ensuring that MLLMs are effective in diverse real-world applications and can adapt to various practical scenarios.

\subsection{Conversation Abilities:}
Some benchmarks focus on evaluating MLLMs' performance in conversations, specifically assessing how well these models handle long contexts and follow complex instructions accurately. Such evaluations are crucial for ensuring that MLLMs can engage effectively in diverse dialogues and deliver reliable performance in real-world applications.

\textbf{Long-context Capabilities:}
Due to the window length limitations inherent in MLLMs' architectures, evaluating their ability to handle long contexts is challenging. This involves assessing whether MLLMs can maintain accurate recall and effective understanding as the amount of contextual information increases.
MileBench \cite{song2024MileBench} and MMNeedle \cite{wang2024MMNeedle} explored the long-context recall abilities of MLLMs using needle-in-a-haystack (NIAH) method and image retrieval tasks. However, evaluating the long-context understanding of MLLMs in videos remained a significant challenge. Therefore, Zhou et al. \cite{zhou2024MLVU} proposed MLVU, which was developed using long videos of diversified lengths.
% MileBench\cite{song2024MileBench} and MMNeedle\cite{wang2024MMNeedle} explored the long-context recall abilities of MLLMs, using needle-in-a-haystack(NIAH) and image retrieval tasks. However, it remains a great challenge to evaluate the MLLMs’ long-context understanding in videos. Therefore, Zhou et al.\cite{zhou2024MLVU} proposed MLVU, which was created based on long videos of diversified lengths.

\textbf{Instruction Adherence:}
Instruction adherence requires that MLLMs execute complicated instructions. This involved not only recognizing the content of the instructions but also meticulously executing the detailed demands without deviation. Wang et al.\cite{wang2023DEMON} built a benchmark called Demon for demonstrative instruction understanding. However, it focused only on demonstrative instruction following and ignored other flexible instruction scenarios. 
To explore how MLLMs performed on broader, open-ended prompts, Bitton et al.\cite{bitton2023VisIT-Bench} created VisIT-Bench to cover a wide array of 'instruction families' that resembled real-world user behavior.
% In exploring how MLLMs performed on broader, open-ended prompts , Bitton et al.\cite{bitton2023VisIT-Bench} created VisIT-Bench to cover a wide array of “instruction families.”that resembled real-world user behavior. 
Although these benchmarks evaluated the basic instruction-following capabilities of MLLMs, the ability of MLLMs to adapt to new instructions while incorporating both old and new ones remained unclear. Therefore, Chen et al.\cite{chen2024CoIN} presented a benchmark named Continual Instruction Tuning (CoIN) to assess MLLMs in a sequential instruction tuning paradigm. 
To better measure MLLM adherence to instructions, MIA-Bench\cite{qian2024MIA-Bench} was introduced to test how well MLLMs follow layered instructions and generate accurate responses matching specific patterns.
% For a more precise measurement of MLLM adherence to instructions, MIA-Bench\cite{qian2024MIA-Bench} was proposed to challenge the compliance of MLLMs with layered instructions in generating accurate responses that satisfied specific requested patterns.
% Instruction adherence requires that MLLMs can execute complicated instructions. This involves not only recognizing the content of the instructions, but also meticulously executing the detailed demands without deviation. Wang et al.\cite{wang2023DEMON} built a benchmark called Demon for demonstrative instruction understanding. However, it only focused on demonstrative instruction following and ignored other more flexible instruction scenarios. In view of exploring how MLLMs performed on broader, open-ended queries that resemble real-world user behavior, Bitton et al.\cite{bitton2023VisIT-Bench} created VisIT-Bench to cover a wide array of “instruction families”. Although these benchmarks can evaluate the basic instruction-following capabilities of MLLMs, MLLMs' ability to adapt to new instructions while incorporating both old and new ones remained unclear. Therefore, Chen et al.\cite{chen2024CoIN} presented a benchmark named Continual Instruction Tuning (CoIN), to assess existing MLLMs in sequential instruction tuning paradigm. For a more precise measurement of MLLM adherence to instructions, MIA-Bench\cite{qian2024MIA-Bench} was proposed to challenge the MLLMs’ compliance with layered instructions in generating accurate responses that satisfy specific requested patterns.

\subsection{Hallucination}
Hallucination refers to information in LVLMs' responses that does not accurately reflect the visual input, which poses potential risks of substantial consequences.
% Hallucination refers to the information of LVLMs’ responses that does not exist in the visual input, which poses potential risks of substantial consequences.

To measure object hallucination, Rohrbach et al.\cite{rohrbach2019CHAIR} proposed CHAIR (Caption Hallucination Assessment with Image Relevance), which assessed captioned objects that were actually present in an image. However, CHAIR was unstable and required complex human-crafted parsing rules for exact matching. Alternatively, POPE\cite{li2023POPE} converted hallucination into a binary classification problem, but it required the input questions to follow specific templates, such as 'Is there a/an <object> in the image?'. In comparison, GAVIE\cite{liu2024GAVIE} can evaluate model hallucination in an open-ended manner without requiring ground-truth answers or pre-designed instruction formats. However, it still focused on evaluating object hallucinations, neglecting other types of hallucinations in MLLMs.
M-HalDetect\cite{gunjal2024M-HalDetect} and MMHAL-BENCH \cite{sun2023MMHAL-BENCH} extended the scope of previous works by not only considering hallucinations related to objects but also addressing other categories such as object attributes and spatial relations. Moreover, Chen et al.\cite{chen2024MHaluBench} presented MHaluBench, a meta-evaluation benchmark that encompassed various hallucination categories and multimodal tasks. Additionally, Zhang et al.\cite{zhang2024MRHal-Bench} introduced MRHalBench to evaluate hallucinations in multi-round dialogues. To assess MLLMs' hallucination in video understanding, Wang et al.\cite{wang2024VideoHallucer} introduced VideoHallucer, the first comprehensive benchmark for hallucination detection in videos.
% M-HalDetect\cite{wang2023M-HalDetect} and MMHAL-BENCH\cite{sun2023MMHAL-BENCH} extended the scope of the previous works by not only considering hallucinations on objects, but also on other categories such as object attribute and Spatial relation. Moreover, Chen et al.\cite{chen2024MHaluBench} presented MHaluBench, a meta-evaluation benchmark that encompassed various hallucination categories and multimodal tasks. 
% Additionally, Zhang et al.\cite{zhang2024MRHal-Bench} introduced MRHalBench to evaluate hallucinations in multi-round dialogues.
% To assessing MLLM's hallucination in video understanding, Wang et al.\cite{wang2024VideoHallucer} introduced VideoHallucer, the first comprehensive benchmark for hallucination detection in videos.

Some benchmarks aimed to explore more cost-effective and feasible methods for evaluating hallucinations. Wang et al.\cite{wang2023HaELM} argued that hallucinations measured using object-based evaluations like POPE merely exploited the judgment bias present in MLLMs, rather than reflecting their actual hallucinations. They proposed Hallucination Evaluation based on Large Language Models (HaELM), the first to utilize LLMs for hallucination evaluation within MLLMs. Considering that reliance on LLMs resulted in significant costs, Wang et al.\cite{wang2024AMBER} proposed an LLM-free multi-dimensional benchmark, AMBER. AMBER provided comprehensive coverage of evaluations for various types of hallucinations and offered detailed annotations to support an LLM-free evaluation pipeline. Despite that, many of the visual hallucination (VH) instances in these benchmarks came from existing datasets, which resulted in a biased understanding of MLLMs’ performance due to the limited diversity of such VH instances. Therefore, Huang et al.\cite{huang2024VHTest} proposed generating diverse VH instances using a text-to-image generative model. Based on this method, they collected a benchmark dataset called VHTest. Additionally, to eliminate the need for costly data annotation and minimize the risk of training data contamination, Cao et al.\cite{cao2024MMECeption} proposed an annotation-free evaluation method that required only unimodal data to measure inter-modality semantic coherence and inversely assessed MLLMs’ tendency to hallucinate.
% Some benchmarks aimed to explore more cost-effective and feasible methods for evaluating hallucinations. Wang et al.\cite{wang2023HaELM} advocated that the hallucinations measured object-based evaluation like POPE merely exploit the judgment bias present in MLLMs, rather than reflecting their hallucination. They proposed Hallucination Evaluation based on Large Language Models(HaELM), the first to
% utilize LLM for hallucination evaluation within MLLMs. 
% Considering that the reliance on LLMs resulted in a huge cost, Wang et al.\cite{wang2024AMBER} proposed an LLM-free multi-dimensional benchmark AMBER. AMBER provided comprehensive coverage of evaluations for various types of hallucination, and offered detailed annotations that help to achieve an LLM-free evaluation pipeline. 
% However, 这些benchmark中的visual hallucination(VH) instances
% 大多数都来自于已经存在的数据集, which resulted in biased understanding of MLLMs’ performance under VH due to limited diversity of such VH instances. Therefore, Huang\cite{huang2024VHTest} proposed to generate a diverse set of VH instances using a text-to-image generative model, and collected a benchmark dataset based on this method. Besides,
% in order to eliminate the need for costly data annotation, minimizes the risk of training data contamination, Cao et al.\cite{cao2024MMECeption} proposed an annotation-free evaluation method that required only unimodal data to measure inter-modality semantic coherence and inversely assessed MLLMs’ tendency to hallucinate. 

There are also some works that explored the causes and mechanisms of hallucination. Observing that MLLMs' strong language bias often overshadowed visual information, leading to an overreliance on language priors rather than visual context, HallusionBench\cite{guan2024HallusionBench} was proposed to focus on diagnosing both the visual illusion and knowledge hallucination of MLLMs. Cui et al.\cite{cui2023Bingo} constructed a benchmark called Bingo, which systematically categorized and analyzed the reasons behind the occurrence of hallucinations. Moreover, Han et al.\cite{han2024CorrelationQA} identified a typical class of inputs that baffled MLLMs: images that were highly relevant but inconsistent with answers, causing MLLMs to suffer from hallucination. To quantify this effect, they proposed CorrelationQA, the first benchmark that assessed the hallucination level given spurious images.
% There are also some works that explored the causes and mechanisms of hallucination. Observing that MLLMs strong language bias often overshadows visual information, leading to an overreliance on language priors rather than the visual context, HallusionBench\cite{guan2024HallusionBench} was proposed to focus on diagnosing both the visual illusion and knowledge hallucination of MLLMs. Cui et al.\cite{cui2023Bingo} constructed a benchmark called Bingo, while systematically categorized and analyzed the reasons behind the occurrence of hallucinations. 
% Moreover, Han et al.\cite{han2024CorrelationQA} identified a typical class of inputs that baffled MLLMs, which consisted of images that were highly relevant but inconsistent with answers, causing MLLMs to suffer from hallucination. To quantify this effect, they proposed CorrelationQA, the first benchmark that assessed the hallucination level given spurious images.

\subsection{Trustworthiness}
Evaluating the trustworthiness of multimodal large language models (MLLMs) encompasses various aspects, including accuracy, consistency across different scenarios, and safety in handling sensitive content. This section focuses on benchmarks that assess MLLMs specifically in terms of robustness and safety. Robustness examines how well the model performs with diverse or unexpected inputs, ensuring reliable outputs across various conditions. Safety evaluates the model's capacity to avoid generating harmful or inappropriate content, thereby protecting users from potential risks. These dimensions are essential for confirming that MLLMs are dependable and safe for real-world applications.

\textbf{Robustness:}
Robustness evaluation in MLLMs involves assessing how well the models handle input disturbances and maintain consistent performance across different data distributions and tasks. Some studies tested the robustness of MLLMs by examining their performance across various input distributions. For instance, to examine MLLMs' visual capability in terms of visual diversity, Cai et al.\cite{cai2023benchlmm} proposed BenchLMM, a benchmark that assessed MLLMs in three distinct styles of distribution shifts: artistic style, sensor style, and application style. In terms of prompt diversity, MLLMs might generate erroneous responses when faced with deceptive or inappropriate instructions. To quantitatively assess this vulnerability, MMR \cite{liu2024MMR} and MAD-Bench \cite{qian2024MAD-Bench} were constructed to comprehensively evaluate MLLMs' capability to resist deceiving or misleading information in the prompt. 
Additionally, some incorrect responses from MLLMs stem from their difficulty in understanding what they can and cannot perceive in images, a capability known as self-awareness in perception, which undermines their robustness.
% Additionally, some incorrect responses from MLLMs arose from their difficulty in understanding what they can and cannot perceive in images, a capability referred to as self-awareness in perception, which undermines their robustness.
Therefore, MM-SAP \cite{wang2024MM-SAP} and VQAv2-IDK \cite{cha2024VQAv2-IDK} were developed to assess MLLMs' understanding of self-awareness, particularly in scenarios where the correct response is 'I Don’t Know'.
Ye et al.\cite{ye2024MM-SpuBench} explored the issue of spurious bias, finding that MLLMs often relied on spurious correlations between non-essential input attributes and target variables for their predictions. To better understand this problem, they introduced MM-SPUBENCH, a comprehensive benchmark designed to evaluate MLLMs' dependence on nine distinct categories of spurious correlations.
% Robustness evaluation in MLLMs involves assessing how well the models handle input disturbances and maintain consistent performance across different data distributions and tasks.
% Some studies test the robustness of MLLMs by examining their performance across different input distributions. 
% For instance, to examine MLMMs' visual capability in terms of visual diversity, Cai et al.\cite{cai2023benchlmm} proposed BenchLMM, a benchmark that assessed MLMMs in three distinct styles of distribution shifts: artistic style, sensor style, and application style.
% In terms of prompt diversity, MLLMs may generate erroneous responses when faced with deceptive or inappropriate instructions. To quantitatively assess this vulnerability, MMR\cite{liu2024MMR} and MAD-Bench\cite{qian2024MAD-Bench} were constructed to comprehensively evaluate MLLMs on their capability to resist deceiving or misleading information in the prompt.
% Besides, some incorrect responses from MLLMs arise from their difficulty in understanding what they can and cannot perceive in images, a capability we refer to as self-awareness in perception, which undermines their robustness.
% Therefore, MM-SAP\cite{wang2024MM-SAP} and VQAv2-IDK\cite{cha2024VQAv2-IDK} were developed to assess MLLMs' understanding of self-awareness, particularly in scenarios where the correct response is "I Don’t Know."

\textbf{Safety:}
Some works primarily test the safety of MLLMs, ensuring they refrain from producing content that poses ethical risks or causes social harm. Driven by the observation that MLLMs tended to respond to malicious questions when a query-relevant image was presented in the dialogue, Liu et al.\cite{liu2024MM-SafetyBench} constructed a safety-measurement benchmark called MM-SafetyBench, which encompassed a wide range of unsafe scenarios.
Meanwhile, by designing specific inputs, jailbreak attacks could induce the model to generate harmful content that might violate human values. To explore this issue, JailBreakV\cite{luo2024JailBreakV-28K} was developed to assess various jailbreak techniques on MLLMs.
% Besides, by designing special inputs, jailbreak attacks can induce the model to provide harmful content that may violate human values.  To explore this issue, JailBreakV\cite{luo2024JailBreakV-28K} was designed to assess various jailbreak techniques to MLLMs. 
Besides, MLLMs were vulnerable to data extraction and privacy leaks during inference, posing risks in privacy-sensitive applications. Therefore, it was crucial to empower privacy information with the "right to be forgotten" through machine unlearning, which removed private or sensitive information from models. To address this issue, Li et al.\cite{li2024MMUBench} established MMUBench to evaluate the efficacy, generality, specificity, fluency, and diversity of machine unlearning methods in MLLMs.
% Meanwhile, MLLMs are vulnerable to data extraction and privacy leaks during inference, posing risks in privacy-sensitive applications. Therefore, it's crucial to empower MLLMs with the "right to be forgotten" through Machine Unlearning, which removes private or sensitive information from models. In this issue, Li et al.\cite{li2024MMUBench} established MMUBench to evaluate the efficacy, generality, specificity, fluency and diversity of machine unlearning methods in MLLMs. 
Moreover, Shi et al.\cite{shi2024SHIELD} introduced SHIELD, a benchmark designed to evaluate the effectiveness of MLLMs in addressing various challenges within the domain of facial security, including face anti-spoofing and face forgery detection.
% Additionally, Shi et al. \cite{shi2024SHIELD} introduced a benchmark called SHIELD, which evaluated the effectiveness of MLLMs in addressing various challenges within the domain of face security, including both face anti-spoofing and face forgery detection.
However, these benchmarks typically examined only one or a few aspects of safety and trustworthiness, lacking a comprehensive evaluation. To address this issue, MultiTrust \cite{zhang2024MultiTrust} and RTVLM \cite{li2024RTVLM} were proposed as comprehensive and unified benchmarks. Both evaluated the safety capabilities of MLLMs across diverse dimensions and tasks.
% However, these benchmarks typically examined only one or a few aspects of safety and trustworthiness. To develop a holistic and standardized evaluation benchmark, MultiTrust \cite{zhang2024MultiTrust} and RTVLM \cite{li2024RTVLM} were proposed as comprehensive and unified benchmarks. Both evaluated the safety capabilities of MLLMs across diverse dimensions and tasks.
% However, these benchmarks typically examined one
% or a few aspects of safety and  trustworthiness. In order to develop holistic and standardized evaluation
% benchmark, MultiTrust\cite{zhang2024MultiTrust} and RTVLM\cite{li2024RTVLM} were proposed as comprehensive and unified benchmarks. Both of them evaluate
% the capabilities of MLLMs in safety across diverse dimensions and tasks.

\section{Other Modalities}
\label{sec:om}

Beyond the image modality, other modalities such as video, audio, and 3D point clouds also contain rich and diverse information reflective of the real world. These modalities provide critical context and enhance the ability to understand complex scenarios. Evaluating MLLMs across these varied modalities is crucial for developing robust and versatile models capable of handling a wide range of complex real-world tasks. This section provides an overview of benchmarks designed to assess MLLMs across these different modalities, aiming to highlight their strengths and identify areas for further improvement.
\subsection{Videos}
Compared to images, the video modality features temporal dynamics and sequential context, involving changes over time and interactions between frames. Therefore, evaluating MLLMs on video-related aspects, such as temporal coherence and action understanding, is crucial for assessing their performance in understanding and interpreting video content.

\textbf{Temporal Perception:}
Temporal perception is a fundamental distinction between video-centered and image-centered applications.
TimeIT \cite{Ren2023TimeIT} was conducted to assess the temporal understanding capabilities of MLLMs. However, TimeIT neglected the distinction between various temporal aspects. To address this issue, MVBench \cite{li2024MVBench}, Perception Test \cite{pătrăucean2023PerceptionTest}, VilMA \cite{kesen2023ViLMA}, and VITATECS \cite{li2023VITATECS} introduced a range of fine-grained temporal aspects, enabling a more comprehensive and nuanced evaluation of temporal perception capabilities.
% TimeIT\cite{Ren2023TimeIT} was conducted to stimulate the  temporal understanding capability of MLLMs. However, TimeIT neglected the distinction between various temporal aspects. 
% To tackle this issue, MVBench\cite{li2024MVBench}, Perception Test\cite{pătrăucean2023PerceptionTest}, VilMA\cite{kesen2023ViLMA}, and VITATECS\cite{li2023VITATECS} introduced a range of fine-grained temporal aspects, enabling a more comprehensive and nuanced evaluation of temporal perception capabilities. 
However, MVBench, VITATECS, and VilMA were limited to single-task formats, and Perception Test was constrained to indoor videos, making these benchmarks less suitable for evaluating MLLMs. In response to these issues, Liu et al. \cite{liu2024TempCompass} proposed TempCompass, a benchmark designed to evaluate the temporal perception abilities of MLLMs across various temporal aspects, different task formats, and diverse types of videos.
% However, MVBench, VITATECS, and VilMA were limited to single task formats, and Perception Test was constrained to indoor videos, making these benchmarks less suitable for evaluating MLLMs. In response to these issues, Liu et al.\cite{liu2024TempCompass} proposed TempCompass, a benchmark designed to evaluate the temporal perception abilities of MLLMs across various temporal aspects, different task formats, and diverse types of videos.
Moreover, focusing on the causal relationships between concrete actions and their effects, Nguyen et al. \cite{nguyen2024OSCaR} proposed OsCaR, a benchmark that included various tasks for understanding causal temporal states.
% Moreover, focusing on the causal relationships between concrete actions and their effects, Nguyen et al.\cite{nguyen2024OSCaR} proposed OsCaR, a benchmark that included various tasks for understanding causal temporal states.
However, the videos in these benchmarks typically had clear temporal structures and scene semantics, whereas activities of daily living often involve temporal unstructuredness, with diverse actions occurring concurrently. To address this, ADLMCQ \cite{chakraborty2024ADLMCQ} was introduced to evaluate MLLMs on tasks related to daily living activities.
% However, the videos in these benchmarks always had evident temporal structures and scene semantics, while activities of daily living were characterized by temporal unstructuredness, where diverse actions might unfold concurrently. To address this, ADLMCQ\cite{chakraborty2024ADLMCQ} was proposed to evaluate MLLMs on activities of daily living tasks.

\textbf{Long Video Understanding:}
Video MLLMs often use key frame extraction for video understanding tasks, which presents challenges as video length increases.
Videos in most benchmarks had clear length limitations, which were insufficient to reflect MLLMs’ long-video understanding capabilities. Some works, like Egoschema\cite{mangalam2023EgoSchema} and MovieChat-1k\cite{song2024MovieChat}, collected long videos and created questions based on them. MovieChat-1k was a benchmark for long video understanding tasks, but many of the questions in MovieChat-1k targeted exact time segments, which degraded the tasks to short-video problems. EgoSchema presented video reasoning tasks using first-person footage. However, it only focused on a few aspects of long videos, rather than offering a comprehensive analysis of long-video understanding. To address these issues, MLVU\cite{zhou2024MLVU} was developed as a comprehensive benchmark with various task categories to evaluate MLLMs’ capabilities in understanding long videos. Additionally, Du et al.\cite{du2024Event-Bench} introduced Event-Bench, an event-oriented long-video understanding benchmark that used diverse videos to evaluate MLLMs’ ability to understand complex event narratives.
% Videos in most benchmarks are no more than five minutes, which are insufficient to reflect the MLLMs’ long-video understanding capabilities. Some works like Egoschema\cite{mangalam2023EgoSchema} and MovieChat-1k\cite{song2024MovieChat} collect long videos and create questions based on them. MovieChat-1k was a benchmark for long video understanding tasks, but many of questions in MovieChat-1k targeted on exact time segments which degraded the tasks to short-video problems. EgoSchema presented video reasoning tasks using first-person footage. However, it only focus a few aspect of MLLMs, rather than offering a comprehensive analysis of long video understanding. To address these issues, Zhou et al.\cite{zhou2024MLVU} developed a comprehensive benchmark with 
% various task categories to effectively evaluate MLLMs’ capabilities in understanding long videos.
% Additionally, Du et al.\cite{du2024Event-Bench} presented an event-oriented long video understanding benchmark with diverse videos to comprehensively evaluate MLLM’s ability to understand
% complex event narratives.

\textbf{Comprehensive Evaluation:}
Comprehensive evaluation of video MLLMs focuses on their overall abilities, including video comprehension, scene recognition, temporal reasoning, and various other related tasks.
With the development of video MLLMs, several works collected extensive video datasets to comprehensively evaluate MLLMs, such as Video-Bench \cite{ning2023Video-Bench} and AutoEval-Video \cite{chen2024AutoEval-Video}.
% With the development of video MLLMs, several works collected diverse videos to comprehensively evaluate MLLMs, such as Video-Bench \cite{ning2023Video-Bench} and AutoEval-Video \cite{chen2024AutoEval-Video}. 
However, they were limited by a lack of diversity in video types and insufficient coverage of temporal dynamics. To create a more comprehensive and high-quality assessment of MLLM in video, Fu et al.\cite{fu2024Video-MME} proposed Video-MME, which offered more diverse video types, broader temporal coverage, and higher quality annotations. Nevertheless, Video-MME employed a multiple-choice question-and-answer format for straightforward and flexible assessment, but this format overlooked the quality and richness of free-form expressions generated by MLLMs. To address this issue, Fang et al.\cite{fang2024MMBench-Video} introduced MMBench-Video, a comprehensive benchmark that consisted of free-form questions mirroring practical use cases.
Additionally, due to the ability of videos to capture rich representations of real-world dynamics and causalities, some works evaluated MLLMs' understanding of videos through the lens of "world models," which required MLLMs to interpret and reason about complex real-world dynamics. MMWorld \cite{he2024MMWorld} and WorldNet \cite{ge2024WorldNet-Crafted} were designed to rigorously evaluate the capabilities of MLLMs in world modeling via video understanding. MMWorld spanned a broad spectrum of disciplines, featuring a diverse array of question types for multi-faceted reasoning, while WorldNet encompassed millions of samples across a wide range of real-world scenarios and tasks.
% With the development of video MLLMs, several works collect diverse videos to evaluate MLLMs comprehensively, such as Video-Bench\cite{ning2023Video-Bench} and AutoEval-Video\cite{chen2024AutoEval-Video}). 
% However, existing benchmarks are still limited by a lack of diversity in video types and insufficient coverage of temporal dynamics. 
% To create a more comprehensive and high-quality assessment of MLLM performance in video, Fu et al. \cite{fu2024Video-MME} proposed Video-MME, which offers more diverse video types, more extensive temporal coverage, a broader range of data modalities, and higher-quality annotations.

% Besides, due to the characteristic of videos in capturing rich representations of real-world dynamics and causalities, some works have evaluated MLLMs' ability to understand videos through the perspective of "world models," which require MLLMs to interpret and reason about complex real-world dynamics.
% MMWorld\cite{he2024MMWorld} and WorldNet\cite{ge2024WorldNet-Crafted} were designed to rigorously evaluate the capabilities
% of MLLMs in world modeling through the realm of video understanding. MMWorld spans a broad spectrum of disciplines, featuring a rich array of question types for multi-faceted reasoning. while WorldNet encompasses millions of samples from a broad spectrum of real-world scenarios and tasks 
\subsection{Audio}
MLLMs that jointly process audio and language hold great promise for audio understanding. 
Dynamic-SUPERB \cite{huang2024Dynamic-SUPERB} was a benchmark that assessed MLLMs’ ability to follow instructions in the audio domain. Nevertheless, Dynamic-SUPERB focused solely on human speech processing, ignoring other types of audio.
% Dynamic-SUPERB \cite{huang2024Dynamic-SUPERB} was a benchmark that assessed MLLMs’ ability to follow instructions in the audio domain. Nevertheless, Dynamic-SUPERB only focused on human speech processing.
To measure the understanding abilities of MLLMs in music, Weck et al. \cite{weck2024MuChoMusic} presented MuChoMusic, the first benchmark for evaluating music understanding in audio MLLMs.
% To measure the understanding abilities of MLLMs
% in music, Weck et al.\cite{weck2024MuChoMusic}presented MuChoMusic, the first benchmark for evaluating music understanding in Audio LLMs.
In order to offer comprehensive coverage of audio signals, including human speech, natural sounds, and music, Yang et al. \cite{yang2024AIR-Bench} presented AIR-Bench, a comprehensive benchmark designed to evaluate MLLMs' ability to comprehend various audio signals and interact according to instructions.
% In view of offering comprehensive coverage of audio signals, Yang et al.\cite{yang2024AIR-Bench} presented AIR-Bench, the first comprehensive benchmark designed to evaluate the ability of MLLMs to comprehend various audio signals and to interact following instructions.

\subsection{3D Scenes}
3D scenes offer a significant advantage over 2D images by providing accurate spatial relationships, depths, and occlusions, which are essential for interpreting complex environments. For MLLMs, enhanced 3D scene perception enables more precise responses in applications such as navigation, augmented reality, and complex spatial reasoning tasks.
ScanQA \cite{azuma2022ScanQA} was proposed for 3D visual question answering, requiring models to answer given questions based on point clouds. It was formatted as an attribute classification task, which made it difficult to assess MLLMs' detailed understanding of 3D scenes. 
LAMM\cite{yin2023LAMM} transformed the classification task into a multiple-choice problem and added two tasks beyond 3D visual question answering: 3D object detection and visual grounding.
% ScanQA\cite{azuma2022ScanQA} was proposed for 3D visual question answering, requiring models to answer given questions based on point clouds. It was formatted as an attribute classification task, which made it difficult to assess MLLMs' detailed understanding of 3D scenes. 
% LAMM transformed the classification task into a multiple-choice problem, and added two tasks beyond 3D visual question answering: 3D object detection and visual grounding.

{
\tiny
\setlength{\tabcolsep}{2pt}
\begin{center}
\begin{longtable}{|c|c|c|c|c|c|}
% \centering
\caption{Summary of existing MLLM evaluations and benchmarks. This table provides an overview of the key attributes of the benchmarks introduced in this paper. The "Focus" describes the primary features and focal points of each benchmark; The "Answer Annotation" denotes the size of the dataset for each benchmark;The "Answer Type" categorizes the format of responses: T/F refers to True or False, Y/N denotes Yes or No, MQA indicates multiple-choice answers, restrictive text refers to text with specific format constraints, and open indicates open-ended responses. "Evaluation" specifies the evaluation methods employed;The "Models" column indicates the number of models evaluated in each benchmark.}
\label{tb-benchmarks} 
%\resizebox{\textwidth}{!}{
%\begin{tabular}{|l|c|c|c|c|c|}
\\
\hline
\textbf{Benchmark} & \textbf{Focus} & \textbf{Answer Annotation} & \textbf{Answer Type} & \textbf{HGEvaluation} & \textbf{Models} \\ \hline
\endfirsthead

\hline
\textbf{Benchmark} & \textbf{Focus} & \textbf{Answer Annotation} & \textbf{Answer Type} & \textbf{HGEvaluation} & \textbf{Models}\\
\hline
\endhead

\hline
\endfoot

\hline
\endlastfoot
% Flicker30k https://arxiv.org/abs/1505.04870
%KB-VQA:https://arxiv.org/pdf/1511.02570
%Visual 7w https://arxiv.org/abs/1511.03416
%FVQA:https://arxiv.org/pdf/1606.05433
%VQAv2:https://arxiv.org/pdf/1612.00837
%CLEVR:https://arxiv.org/pdf/1612.06890
%VizWiz:https://arxiv.org/abs/1802.08218
%Chair:https://arxiv.org/abs/1809.02156
%GQA:https://arxiv.org/pdf/1902.09506
%RAVEN:https://arxiv.org/abs/1903.02741
% TextVQA:https://arxiv.org/abs/1904.08920
%OK-VQA:https://arxiv.org/pdf/1906.00067
% TextCaps:https://arxiv.org/abs/2003.12462
%SPDocVQA:https://arxiv.org/abs/2007.00398
%InfographicVQA:https://arxiv.org/abs/2104.12756
%Geometry3K:https://arxiv.org/abs/2105.04165
% ScanQA:https://arxiv.org/abs/2112.10482
%ChartQA:https://arxiv.org/pdf/2203.10244
% AVQA \2022.00000
% MusicAVQA https://arxiv.org/abs/2203.14072
%Winoground:https://arxiv.org/abs/2204.03162
%VSR https://arxiv.org/abs/2205.00363
VSR\cite{VSR}&visual spatial reasoning&10972&T/F&N/A&4 \\ \hline
%A-OKVQA:https://arxiv.org/pdf/2206.01718
%VALSE:http://dx.doi.org/10.18653/v1/2022.acl-long.567
VALSE\cite{VALSE}&visio-linguistic grounding capabilities&6795&A/Bs&N/A&6\\ \hline
%VL-CheckList:https://arxiv.org/abs/2207.00221
VL-CheckList\cite{zhao2023VL-CheckList}&image-text matching&&A/B&N/A&7\\ \hline
%ScienceQA:https://arxiv.org/pdf/2209.09513
%Aro:https://arxiv.org/abs/2210.01936
ARO\cite{yuksekgonul2023ARO}&composition and order understanding&&MQA&N/A&4\\ \hline
%MPDocVQA:https://arxiv.org/abs/2212.05935
%Eqben:https://arxiv.org/abs/2303.14465
Eqben\cite{wang2023Eqben}&video-text matching&250k&A/B&N/A&9\\ \hline
%LLaVA-Bench:https://arxiv.org/pdf/2304.08485
LLaVA-Bench\cite{liu2023LLaVA-Bench} & visual
instruction following & 54 & open & GPT & 4 \\ \hline
%OwlEval:https://arxiv.org/abs/2304.14178
OwlEval\cite{ye2024OwlEval}& comprehensive evaluation & 82 & open & Human & 5 \\ \hline
% OCRBench:https://arxiv.org/abs/2305.07895
OCRBench\cite{liu2024OCRBench}&OCR capabilities&1000&open&N/A&14\\ \hline
%DUDE:https://arxiv.org/abs/2305.08455
DUDE\cite{vanlandeghem2023DUDE}&understanding
visually-rich documents&11448&open&N/A&6\\ \hline
%POPE:https://arxiv.org/abs/2305.10355
POPE\cite{li2023POPE}&object hallucination&3000&Y/N&N/A&5\\ \hline
%GVT-Bench: https://arxiv.org/abs/2305.12223
% Perception Test:https://arxiv.org/abs/2305.13786
PerceptionTest\cite{pătrăucean2023PerceptionTest}&Memory, Abstraction, Physics, Semantics& 11.6k&MQA, open&N/A&\\ \hline
%NuScenes-QA:https://arxiv.org/abs/2305.14836
NuScenes-QA\cite{qian2024NuScenes-QA}&autonomous driving&460k&open&N/A&9\\ \hline
%CODE:https://arxiv.org/abs/2305.18279
CODE\cite{zang2023CODE} & contextual object detection & 10346 &  open & - & 2 \\ \hline
%M3Exam:https://arxiv.org/abs/2306.05179
M3Exam\cite{zhang2023M3Exam}& official human exam questions&12317&MQA&N/A&7\\ \hline
%LAMM:https://arxiv.org/abs/2306.06687
LAMM\cite{yin2023LAMM} &  instruction tuning & 62,439 & open & GPT & - \\ \hline
% LAMM:https://arxiv.org/abs/2306.06887
LAMM\cite{yin2023LAMM}&multi-modal instruction
tuning&62439&open&GPT&4\\ \hline
%MME:https://arxiv.org/abs/2306.13394
MME\cite{MME} & perception and cognition & 1.5k & Y/N & N/A & 10 \\ \hline
% GAVIE:https://arxiv.org/abs/2306.14565
%MMBench:https://arxiv.org/pdf/2307.06281
MMBench\cite{liu2024MMBench} & objective evaluation & 3217 & open & GPT & 14 \\ \hline
%SeedBench:https://arxiv.org/abs/2307.16125
%MovieChat:https://arxiv.org/pdf/2307.16449
MovieChat\cite{song2024MovieChat}&long video understanding&14k&open&GPT&6\\ \hline
%MM-Vet:https://arxiv.org/abs/2308.02490
MM-Vet\cite{yu2023MM-Vet} & integrated capabilities & 205 & open & GPT & 9 \\ \hline
%SciGraphQA:https://arxiv.org/abs/2308.03349
SciGraphQA\cite{li2023SciGraphQA}& scientific question-answering&657k&open&GPT&5\\ \hline
% DEMON:https://arxiv.org/abs/2308.04152
DEMON\cite{wang2023DEMON}&demonstrative instruction understanding& 477.72K&open&N/A&9\\ \hline
% M-HalDetect:https://arxiv.org/abs/2308.06394
M-HalDetect\cite{gunjal2024M-HalDetect}&hallucination detection&16K&open&human&3\\ \hline
% VisIT-Bench:https://arxiv.org/abs/2308.06595
VisIT-Bench\cite{bitton2023VisIT-Bench}& instruction following&592&open&Human/GPT&14\\ \hline
%EgoSchema:https://arxiv.org/pdf/2308.09126
EgoSchema\cite{mangalam2023EgoSchema}&long video understanding&5063&MQA&N/A&4\\ \hline
%TouchStone:https://arxiv.org/abs/2308.16890
TouchStone\cite{bai2023TouchStone}& open-ended real-world dialogues & 908 & open & GPT & 7 \\ \hline
% Dynamic-SUPERB:https://arxiv.org/abs/2309.09510
%Q-Bench:https://arxiv.org/abs/2309.14181
Q-Bench\cite{wu2024Q-Bench}&low-level visual perception&2990&Y/N,open&N/A&15\\ \hline
% HaELM:arXiv:2308.15126
% MMHAL-BENCH:https://arxiv.org/abs/2309.14525
MMHAL-BENCH\cite{sun2023MMHAL-BENCH}&penalizing hallucinations&96&open&GPT&6\\ \hline
%PCA-EVAL:https://arxiv.org/abs/2310.02071
PCA-EVAL\cite{chen2023PCA-EVAL}&decision-making ability&300&MQA&GPT&6\\ \hline
%MathVista:https://arxiv.org/abs/2310.02255
MathVista\cite{lu2024MathVista}&mathematical tasks&735&MQA,open&GPT&12\\ \hline
%MM-Edit:https://arxiv.org/abs/2310.08475
MMEdit\cite{cheng2024MMEdit}&knowledge
editing&&open&N/A&2\\ \hline
% HallusionBench:https://arxiv.org/abs/2310.14566
HallusionBench\cite{guan2024HallusionBench}& image-context reasoning&1129&open&GPT&15\\ \hline
%What'sUp https://arxiv.org/abs/2310.19785
What'sUp\cite{kamath2023what'sup} & spatial relations & 820 &  text options & N/A  & 8 \\ \hline
%CHEF:https://arxiv.org/abs/2311.02692
ChEF\cite{shi2023ChEF} & Comprehensive Evaluation Framework & - & - & - & 9 \\ \hline
% Bingo:https://arxiv.org/abs/2311.03287
Bingo\cite{cui2023Bingo}&hallucinations related to bias and interference&370&open&GPT&2\\ \hline
%MagnifierBench:https://arxiv.org/abs/2311.04219
MagnifierBench\cite{li2023MagnifierBench} & visual perception of small objects & 283 & MQA, open & Human/GPT & - \\ \hline
% VILMA:https://arxiv.org/abs/2311.07022
ViLMA\cite{kesen2023ViLMA}&visio-linguistic capabilities&4177&caption matching&N/A&12\\ \hline
% AMBER:https://arxiv.org/abs/2311.07397
AMBER\cite{wang2024AMBER}&hallucinations related to attributes and relations&15k&Y/N, open&N/A&9\\ \hline
%MMC-Benchmark:https://arxiv.org/abs/2311.10774
MMC-Benchmark\cite{liu2024MMC-Benchmark}&visual charts understanding&2k&Open,MQA&GPT&6\\ \hline
%Charting New Territories:https://arxiv.org/abs/2311.14656
%AutoEval-Video:https://arxiv.org/abs/2311.14906
AutoEval-Video\cite{chen2024AutoEval-Video}& open ended video question answering&327&open&GPT&11\\ \hline
%Video-Bench:https://arxiv.org/abs/2311.16103
Video-Bench\cite{ning2023Video-Bench}&comprehensive evaluation for video&17036&MQA&N/A&8\\ \hline
%MMMU:https://arxiv.org/pdf/2311.16502
MMMU\cite{yue2024MMMU}&college-level subject knowledge&11.5k&MQA,open&N/A&23\\ \hline
% MVBench:https://arxiv.org/abs/2311.17005
MVBench\cite{li2024MVBench}& video understanding&4000&MQA&N/A&17\\ \hline
%SEED-Bench-2:https://arxiv.org/abs/2311.17092
% VITATECS:https://arxiv.org/abs/2311.17404
VITATECS\cite{li2023VITATECS}&temporal concept underStanding&13k&open&N/A&9\\ \hline
% MM-SafetyBench:https://arxiv.org/abs/2311.17600
MM-SafetyBench\cite{liu2024MM-SafetyBench}&safety-critical evaluations&5040&open&GPT&12\\ \hline
% TimeIT:abs/2312.02051
TimeIT\cite{Ren2023TimeIT}& long video understanding&125k&open&&8\\ \hline
%Grounding-Bench:https://arxiv.org/abs/2312.02949
Grounding-Bench\cite{zhang2023Grounding-Bench}&grounding and
chat capabilities&1000&open&GPT&9\\ \hline
%EgoPlan-Bench:https://arxiv.org/abs/2312.06722
EgoPlan-Bench\cite{chen2024EgoPlan-Bench}& human-level planning&5k&MQA&N/A&28\\ \hline

% M3DBench:https://arxiv.org/abs/2312.10763
M3DBench\cite{li2023M3DBench}&3D instruction following&327k&open&GPT&3\\ \hline

%v* Bench https://arxiv.org/abs/2312.14135
V*Bench\cite{wu2023V*Bench} & grounding detailed
visual information &191&MQA& & 7 \\ \hline

%DriveLM-DATA:https://arxiv.org/abs/2312.14150
DriveLM-DATA\cite{sima2024DriveLM-Data}&autonomous driving&&&&\\ \hline

%ChartBench:https://arxiv.org/abs/2312.15915
ChartBench\cite{xu2024ChartBench}& chart comprehension&600k&open&N/A&21\\ \hline
%TransportationGame:https://arxiv.org/abs/2401.04471
TransportationGames\cite{zhang2024TransportationGames}&transportation&1910&MQA,T/F,open&GPT&16\\ \hline
%MMVP:https://arxiv.org/abs/2401.06209
MMVP\cite{tong2024MMVP}&visual perception&300&MQA& &9\\ \hline
%BenchLMM:arXiv preprint arXiv:2312.02896
BenchLMM\cite{cai2023benchlmm}&robustness&&open&GPT&10\\ \hline
%MM-SAP:https://arxiv.org/abs/2401.07529
MM-SAP\cite{wang2024MM-SAP}&self-awareness in perception for MLLMs&1150&MQA&N/A&14\\ \hline
%AesBench:https://arxiv.org/abs/2401.08276
AesBench\cite{huang2024AesBench}& image aesthetics perception&8400&MQA&GPT&15\\ \hline
%SPEC:https://arxiv.org/abs/2312.0008
SPEC\cite{peng2024SPEC}&fine-grained comprehension&&text options&N/A&5\\ \hline

%Mementos:https://arxiv.org/abs/2401.10529
Mementos\cite{wang2024Mementos}& sequential
image reasoning abilities&4761&free-from&GPT&9\\ \hline
%CMMMU:https://arxiv.org/abs/2401.11944
CMMMU\cite{zhang2024CMMMU}& college-level subject knowledge&12K&open&N/A&11\\ \hline
% RTVLM:https://arxiv.org/abs/2401.12915
RTVLM\cite{li2024RTVLM}& red teaming&5200&open&GPT&10\\ \hline
%MMMU和CMMMU、CMMU都是自建pipeline
%CMMU:https://arxiv.org/pdf/2401.14011
CMMU\cite{he2024CMMU}&knowledge comprehension and reasoning&3603&open&N/A&10\\ \hline
%Mobile-Eval:https://arxiv.org/abs/2401.16158
Mobile-Eval\cite{wang2024Mobile-Eval}&mobile device operations&33&N/A&N/A&\\ \hline
%LHRS-Bench:https://arxiv.org/abs/2402.02544
LHRS-Bench\cite{muhtar2024LHRS-Bench}&remote sensing&690&MQA&N/A&9\\ \hline
%MULTI:https://arxiv.org/abs/2402.03173

% MHaluBench:https://arxiv.org/abs/2402.03190
MHaluBench\cite{chen2024MHaluBench}& multimodal hallucination detection&420&open&N/A&2\\ \hline
% CorrelationQA:https://arxiv.org/abs/2402.03757
CorrelationQA\cite{han2024CorrelationQA}& instinctive bias across different types&7308&open&N/A&9\\ \hline
% SHIELD:https://arxiv.org/abs/2402.04178
SHIELD\cite{shi2024SHIELD}& face spoofing and forgery detection&&T/F, MQA &N/A&2\\ \hline
%SceMQA:https://arxiv.org/pdf/2402.05138
SceMQA\cite{liang2024SceMQA}&core science subjects&1045&open&GPT&6\\ \hline
%Q-Bench+:https://arxiv.org/abs/2402.07116
Q-Bench+\cite{zhang2024Q-Bench+}&low-level visual perception&949&Y/N,open&N/A&24\\ \hline
% AIR-Bench:https://arxiv.org/abs/2402.07729

%VQAv2-IDK:https://arxiv.org/abs/2402.09717
VQAv2-IDK\cite{cha2024VQAv2-IDK}&self-awareness hallucination&20k&open&N/A&3\\ \hline
%Asclepius:https://arxiv.org/abs/2402.11217
Asclepius\cite{wang2024Asclepius}&medicine&3232&MQA&N/A&6\\ \hline
%ChartX:https://arxiv.org/abs/2402.12185
ChartX\cite{xia2024ChartX}&chart reasoning&48k&open&N/A&11\\ \hline
% MCUB:https://arxiv.org/abs/2402.12750
MCUB\cite{chen2024MCUB}&understand inputs from
diverse modalities&&MQA&N/A&5\\ \hline
%MAD-Bench:https://arxiv.org/abs/2402.13220
MAD-Bench\cite{qian2024MAD-Bench}& resist deceiving information in the prompt&1000&open&GPT&19\\ \hline
%CODIS:https://arxiv.org/pdf/2402.13607
CODIS\cite{luo2024CODIS} & context-dependent
visual comprehension & 706 & Close\&Open-ended & Human/ GPT & 14 \\ \hline
%MM-Soc:https://arxiv.org/abs/2402.14154
MM-Soc\cite{jin2024MM-Soc}&social media content&&MQA,open&N/A&10\\ \hline
% VHTest:https://arxiv.org/abs/2402.14683
VHTest\cite{huang2024VHTest}&visual hallucination&1200&Y/N&N/A&8\\ \hline
%Math-Vision:https://arxiv.org/abs/2402.14804
Math-Vision\cite{wang2024Math-Vision}&mathematical reasoning&3k&open&GPT&9\\ \hline
%MIKE:https://arxiv.org/abs/2402.14835
MIKE\cite{li2024MIKE}& knowledge editing&1000 entities&open&N/A&2\\ \hline
%MMEcaption:https://arxiv.org/abs/2402.14973
MMEcaption\cite{cao2024MMECeption}& inter-modality semantic coherence&&N/A&N/A&7\\ \hline
% OSCaR:https://arxiv.org/abs/2402.17128
OSCaR\cite{nguyen2024OSCaR}&object state understanding&14084&open&N/A&8\\ \hline
%CRPE https://arxiv.org/abs/2402.19474
CRPE\cite{wang2024CRPE} & relation comprehension & - & MQA & N/A & 3 \\ \hline
% TempCompass:https://arxiv.org/abs/2403.00476
TempCompass\cite{liu2024TempCompass}&temporal perception&7540&multiple types of tasks&GPT&11\\ \hline
%Henna:https://arxiv.org/abs/2403.01031
Henna\cite{alwajih2024Henna}&Arabic culture&1132&open&GPT&2\\ \hline
%NPHardEval4V:https://arxiv.org/abs/2403.01777
NPHardEval4V\cite{fan2024NPHardEval4V}& reasoning abilities&900&restrictive text&N/A&9\\ \hline
%VLKEB:https://arxiv.org/abs/2403.07350
VLKEB\cite{huang2024VLKEB}&knowledge editing&3174&open&N/A&5\\ \hline
% CoIN:https://arxiv.org/abs/2403.08350
CoIN\cite{chen2024CoIN}& continual instruction tuning&&open&N/A&2 \\ \hline
%VL-ICL Bench:https://arxiv.org/abs/2403.13164
VL-ICL Bench\cite{zong2024VL-ICLBench} &  in-context learning & 1520 & open & N/A & 12 \\ \hline

%MathVerse:https://arxiv.org/abs/2403.14624
MathVerse\cite{MathVerse}& visual math tasks & 15k &open&GPT&17\\ \hline
%visualCoT:https://arxiv.org/abs/2403.16999
VisualCoT\cite{shao2024VisualCoT} & specific local region identification &&free-from&GPT&3\\ \hline

%P2GB:https://arxiv.org/abs/2403.19322
P2GB\cite{chen2024P2GB}&  visual reasoning capabilities& 2130 & MQA & - & 3 \\ \hline
%MDVP-Bench:https://arxiv.org/abs/2403.20271
MDVP-Bench\cite{lin2024MDVP-Bench} & visual prompting research & - & Y/N & GPT & 4  \\ \hline
%MMStar:https://arxiv.org/pdf/2403.20330
MMStar\cite{chen2024MMStar}&vision-indispensable tasks & 1500 &MQA&N/A &16\\ \hline

%M3D:https://arxiv.org/abs/2404.00578
M3D\cite{bai2024M3D}&3D medical tasks&&MQA,open&N/A&2\\ \hline
% JailBreakV-28K:https://arxiv.org/abs/2404.03027
JailBreakV-28K\cite{luo2024JailBreakV-28K}&the robustness of MLLMs against jailbreak attacks&28000&open&LLM&10\\ \hline
%FABABench:https://arxiv.org/abs/2404.05052
FABA-Bench\cite{li2024FABA-Bench}&facial affective behavior analysis&403&open&N/A&5\\ \hline
%Freet-UI:https://arxiv.org/abs/2404.05719
Ferret-UI\cite{you2024Ferret-UI}&understanding of mobile UI screens&69&N/A&N/A&4\\ \hline
%VisualWebBench:https://arxiv.org/abs/2404.05955
VisualWebBench\cite{liu2024VisualWebBench}& understanding and grounding in web scenarios & 1.5k &open&N/A&14\\ \hline
%DesignQA:https://arxiv.org/abs/2404.07917
DesignQA\cite{doris2024DesignQA}&real-world engineering tasks&1451&open&N/A&5\\ \hline
%LavyBench:https://arxiv.org/abs/2404.07922
Lavy\cite{tran2024LaVy}&vietnamese visual language understanding&&&&\\ \hline
%UNIAA:https://arxiv.org/abs/2404.09619
UNIAA\cite{zhou2024UNIAA}&image aesthetic attributes&5354&multiple types of tasks&N/A&13\\ \hline
%CFMM:https://arxiv.org/pdf/2404.12966
CFMM\cite{li2024CFMM} & counterfactual reasoning & 2400 & A/B & N/A & 7 \\ \hline
%MARVEL:https://arxiv.org/abs/2404.13591
MARVEL\cite{jiang2024MARVEL}&abstract visual reasoning&770&MQA&N/A&9\\ \hline
%DesignProbe:https://arxiv.org/abs/2404.14801
DesignProbe\cite{lin2024DesignProbe}&design task&open&GPT&&9\\ \hline

%ImplicitVAE:https://arxiv.org/abs/2404.15592
ImplicitAVE\cite{zou2024ImplicitAVE}&implicit value extraction&1610&restrictive text&N/A&6\\ \hline
%MMT-Bench:https://arxiv.org/abs/2404.16006
MMT-Bench\cite{ying2024MMT-Bench}&general purpose multimodal intelligence & 31k & MQA & GPT & 30\\ \hline

%SEED-Bench-2-Plus:https://arxiv.org/abs/2404.16790
SEED-Bench-2-Plus\cite{li2024SEED-Bench-2-Plus}&text-rich visual comprehension & 2.3k & MQA & N/A &34\\ \hline
%WorldNet:https://arxiv.org/abs/2404.18202

%Milebench:https://arxiv.org/abs/2404.18532
MileBench\cite{song2024MileBench}&long-context capabilities&6440&MQA&N/A&22\\ \hline
%Mile-Bench:https://arxiv.org/abs/2404.18532
MileBench\cite{song2024MileBench}&multimodal long-context capabilities&6440&MQA&N/A&22\\ \hline
%Plot2Code:https://arxiv.org/abs/2405.07990
Plot2Code\cite{wu2024Plot2Code}&multi-modal code tasks&132&open&GPT&14\\ \hline

%SciFiBench:https://arxiv.org/abs/2405.08807
SciFIBench\cite{roberts2024SciFIBench}&scientific figure interpretation&1000&MQA&N/A&26\\ \hline
%SOK-Bench:https://arxiv.org/pdf/2405.09713
SOK-Bench\cite{wang2024SOK-Bench}&video reasoning&44K&MQA&N/A&7\\ \hline

%MRHal-Bench:https://arxiv.org/abs/2405.11165
MRHal-Bench\cite{zhang2024MRHal-Bench}&hallucinations in multi-round dialogues&105&open&GPT&15\\ \hline
%MTVQA:https://arxiv.org/abs/2405.11985
MTVQA\cite{tang2024MTVQA}&multilingual text-rich scenarios&28607&open&N/A&18\\ \hline
% MMUBench:https://arxiv.org/abs/2405.12523
MMUBench\cite{li2024MMUBench}&evaluation of machine unlearning & 1000&open&N/A&4\\ \hline
%M3CoT:https://arxiv.org/abs/2405.16473
M3CoT\cite{chen2024M3CoT}& multi-domain
multi-step chain-of-thought reasoning&11459&MQA&N/A&5\\ \hline
%Video-MME:https://arxiv.org/abs/2405.21075
Video-MME\cite{fu2024Video-MME}& comprehensive evaluation in video analysis&2700&MQA&N/A&12\\ \hline
% SpatialRGPT:https://arxiv.org/abs/2406.01584
SpatialRGPT\cite{cheng2024SpatialRGPT}&grounded spatial reasoning&1406&open&GPT&9\\ \hline
%MLVU:https://arxiv.org/abs/2406.04264
MLVU\cite{zhou2024MLVU}&multi-task long Video Understanding&2593&MQA,open&GPT&20\\ \hline

%MLVU:https://arxiv.org/abs/2406.04264
MLVU\cite{zhou2024MLVU}&multi-task long video understanding&2593&MQA&N/A&20\\ \hline
%M3GIA:https://arxiv.org/abs/2406.05343
M3GIA\cite{song2024M3GIA}&intelligence tests&1800&MQA&N/A&24\\ \hline
%II-Bench:https://arxiv.org/abs/2406.05862
II-Bench\cite{liu2024II-Bench}&higher-order perceptual&1434&MQA&N/A&20\\ \hline
%CVQA:https://arxiv.org/abs/2406.05967
CVQA\cite{romero2024CVQA}&culturally-diverse multilingual VQA&9k&MQA&N/A&8\\ \hline

% MultiTrust:https://arxiv.org/abs/2406.07057
MultiTrust\cite{zhang2024MultiTrust}&the trustworthiness of MLLMs&&open&N/A&21\\ \hline

%MM-NIAH:https://arxiv.org/abs/2406.07230
MM-NIAH\cite{wang2024MM-NIAH}&long multimodal documents&12k&restrictive text&&9\\ \hline

%MMWorld:https://arxiv.org/abs/2406.08407
MMWorld\cite{he2024MMWorld}&multi-discipline, multi-faceted video understanding&6627&MQA, open&GPT&12\\ \hline
%MMRel https://arxiv.org/abs/2406.09121
MMRel\cite{nie2024MMRel} & inter-object relations & 15k & Y/N & N/A & 6 \\ \hline
% ADLMCQ:https://arxiv.org/abs/2406.09390
ADLMCQ\cite{chakraborty2024ADLMCQ}& activities of daily living&&MQA&N/A&5\\ \hline
%MuirBench:https://arxiv.org/abs/2406.09411
MuirBench\cite{wang2024MuirBench}&multi-image understanding&2600&MQA&N/A&20\\ \hline
%Cog-Bench:https://arxiv.org/abs/2406.10424
VCog-Bench\cite{cao2024VCog-Bench} & abstract visual reasoning & 1440& MQA & N/A &16\\ \hline

%MMR:https://arxiv.org/abs/2406.10638
MMR\cite{liu2024MMR}& robustness to leading questions & 600&MQA&N/A&18\\ \hline
%MMNeedle:https://arxiv.org/abs/2406.11230
MMNeedle\cite{wang2024MMNeedle}&long-context capabilities&280K&restrictive text&N/A&11 \\ \hline
%MC-MKE:https://arxiv.org/abs/2406.13219
MC-MKE\cite{MC-MKE}&knowledge editing&2884&open&N/A&2\\ \hline

%GSR-BENCH https://arxiv.org/abs/2406.13246
GSR-BENCH\cite{rajabi2024GSR-BENCH} & spatial relationships & 820 & Template-based generation & N/A & 9 \\ \hline

%Event-Bench:https://arxiv.org/abs/2406.14129
Event-Bench\cite{du2024Event-Bench}&event-oriented long video understanding&2190&MQA&N/A&12\\ \hline
%VideoHallucer:https://arxiv.org/abs/2406.16338
VideoHallucer\cite{wang2024VideoHallucer}&hallucination detection in video&1800&Y/N&N/A&12\\ \hline

%CV-Bench:https://arxiv.org/abs/2406.16860
CV-Bench\cite{tong2024CV-Bench}&vision-centric tasks & 2638 & open & GPT & \\ \hline
%MM-SpuBench:https://arxiv.org/abs/2406.17126
MM-SpuBench\cite{ye2024MM-SpuBench}&spurious
biases in MLLMs&10773&MQA&N/A &15\\ \hline
%Charxiv:https://arxiv.org/abs/2406.18521
Charxiv\cite{wang2024CharXiv}& diverse charts understanding&10k&open&GPT&24\\ \hline
%MMRO:https://arxiv.org/abs/2406.19693
MMRo\cite{li2024MMRo}&robot applications&26175&MQA, open&GPT&13\\ \hline
%Web2Code:https://arxiv.org/abs/2406.20098
Web2Code\cite{yun2024Web2Code}&understanding of the web content&5990&Y/N, open&GPT&4\\ \hline
% MIA-Bench:https://arxiv.org/abs/2407.01509
MIA-Bench\cite{qian2024MIA-Bench}&layered instructions following&400&open&GPT&29\\ \hline

%CRAB:https://arxiv.org/abs/2407.01511
CRAB\cite{xu2024CRAB}& cross-environment tasks&100&N/A&N/A&4\\ \hline

% ScanReason:https://arxiv.org/abs/2407.01525
ScanReason\cite{zhu2024ScanReason}&3D reasoning grounding&10k&bounding box&N/A&11\\ \hline
%MathCheck:https://arxiv.org/abs/2407.08733
MATHCHECK-GEO\cite{zhou2024MATHCHECK}&geometry reasoning&1440&open&GPT&11\\ \hline

%CHOPINLLM:https://arxiv.org/abs/2407.14506
CHOPINLLM\cite{fan2024CHOPINLLM}&diverse charts understanding&6k&open&&\\ \hline

%CompBench:https://arxiv.org/abs/2407.16837

% MuChoMusic:https://arxiv.org/abs/2408.01337
%MMIU:https://arxiv.org/abs/2408.02718
MMIU\cite{meng2024MMIU}&multi-image tasks&11K&MQA&N/A&24\\ \hline
%GMAI-MMBench https://arxiv.org/abs/2408.03361
GMAI-MMBench\cite{chen2024GMAI-MMBench}&medical applications&26K&MQA&GPT&50\\ \hline
%UniBench:https://arxiv.org/abs/2408.04810
%MuCR:https://www.arxiv.org/abs/2408.08105
MuCR\cite{li2024MuCR}&multimodal causal reasoning&400&MQA, open&N/A&17\\ \hline
%SPARK:https://arxiv.org/abs/2408.12114v
SPARK\cite{yu2024SPARK}&multi-vision sensor perception and reasoning&6,248&Y/N, MQA&N/A&10\\ \hline
%MultiMed:https://arxiv.org/abs/2408.12682
MultiMed\cite{mo2024MultiMed}&multimodal and multitask medical understanding&2.56 million&open&domain-specific metrics&\\ \hline
%HR-Bench:https://arxiv.org/abs/2408.15556
HR-Bench\cite{wang2024HR-Bench}&high-resolution
image perception&200&MQA&N/A&13\\ \hline
\end{longtable}
\end{center}
}

Besides, Zhu et al. \cite{zhu2024ScanReason} introduced a benchmark named ScanReason, which required MLLMs to conduct joint reasoning on the question and the 3D environment before predicting the 3D locations of target objects. To evaluate MLLMs' capability to accurately perceive fine-grained spatial relations such as depth, direction, and distance, Cheng et al. \cite{cheng2024SpatialRGPT} proposed SpatialRGPT, which specifically focused on MLLMs' ability to understand 3D spatial concepts like metric distance or size differences between objects.
% Besides, Zhu et al.\cite{zhu2024ScanReason}introduced a benchmark named ScanReason, which required MLLMs to conduct joint reasoning on the question and the 3D environment before predicting the 3D locations of target objects. 
% To evaluate MLLMs' capability of accurately perceiving fine-grained spatial relations such as depth, direction and distance, Cheng et al.\cite{cheng2024SpatialRGPT} proposed SpatialRGPT, which specifically focused on MLLMs' ability to
% understand 3D spatial concepts like metric distance or size differences between objects. 
Meanwhile, a comprehensive evaluation benchmark for accurately assessing the capability of MLLMs on 3D tasks was crucial. In view of this, Li et al. \cite{li2023M3DBench} introduced a comprehensive 3D-centric benchmark called M3DBench, which served as the foundation for developing a versatile and practical general-purpose assistant in real-world 3D environments.
% However, there was still a lack of a comprehensive evaluation benchmark in accurately assessing the capability of MLLMs on 3D tasks. To fill this gap, Li et al.\cite{li2023M3DBench} introduced a comprehensive 3D-centric called M3DBench, serving as the foundation for developing a versatile and practical general-purpose assistant in the real-world 3D environment.

\subsection{Omnimodal}
In the real world, people are surrounded by audio, images, videos, and text messages in their daily lives. These diverse modalities collectively enhance our ability to perceive and understand scenes. However, most benchmarks focus on a single modality, which limits their ability to adapt to real-world environments that involve multiple modalities. Therefore, some works focused on evaluating MLLMs' capabilities in handling multiple modalities simultaneously.
MusicAVQA \cite{li2022MusicAVQA} was constructed to evaluate MLLMs' ability to answer questions regarding visual objects, sounds, and their associations together. However, the visual scenes in MusicAVQA were limited to music performances, where the questions were only about instrument relationships, lacking exploration of more real-life scenarios. In response to this limitation, Yang et al. \cite{AVQA} proposed AVQA, which was designed for audio-visual question answering on general videos of real-life scenarios. Furthermore, to cover a wider range of modal information, Chen et al. \cite{chen2024MCUB} introduced a benchmark called the Multimodal Commonality Understanding Benchmark (MCUB), which included four modalities—image, audio, video, and point cloud. The task of MCUB was to measure the model’s ability to identify commonalities among input entities from diverse modalities but overlooked the various other relationships among different modalities. To achieve general-purpose multimodal intelligence, Ying et al.\cite{ying2024MMT-Bench} present MMT-Bench, a comprehensive benchmark designed to assess MLLMs across massive multimodal tasks, which required expert knowledge and deliberate visual recognition, localization, reasoning, and planning.

\section{Conclusion}

Evaluation carries profound significance and is becoming imperative in the advancement of AGI models. It ensures that the models are not only performing as expected but also meeting the desired standards of accuracy, robustness, and fairness. Through rigorous evaluation, we can identify strengths and weaknesses, guide further improvements, and build trust in the deployment of AI systems in real-world applications. In this study, we provide a comprehensive overview of the evaluation and benchmarks of MLLMs, categorizing them into perception and understanding, cognition and reasoning, specific domains, key capabilities, and other modalities. We aim to enhance the understanding of the current status of MLLMs, elucidate their strengths and limitations, and provide insights into the future progression of MLLMs.
Given the dynamic nature of this field, it's possible that some recent developments may not be fully covered. To address this, we plan to continuously update and enhance the information on our website, incorporating new insights as they emerge.

\newpage
{
    \small
    \bibliographystyle{unsrt}
    \bibliography{reference}

}

\end{document}